\documentclass{article} 
\usepackage{iclr2026_conference,times}

\usepackage{booktabs}
\usepackage{longtable}
\usepackage{graphicx}
\usepackage{multirow}
\usepackage{amsmath}
\usepackage{amsfonts}
\newcommand{\mypara}[1]{\vspace{0.5em}\noindent\textbf{#1}}
\usepackage{algorithm}
\usepackage{algpseudocode} 
\usepackage{marvosym} 

\usepackage{amsmath,amsfonts,bm}

\def\eqref#1{equation~\ref{#1}}

\def\1{\bm{1}}

\DeclareMathAlphabet{\mathsfit}{\encodingdefault}{\sfdefault}{m}{sl}
\SetMathAlphabet{\mathsfit}{bold}{\encodingdefault}{\sfdefault}{bx}{n}

\usepackage{hyperref}
\usepackage{url}
\usepackage{fancyhdr}

\fancypagestyle{myhdr}{
  \fancyhf{}                       
  \fancyhead[L]{\footnotesize Under Review}  
  \fancyfoot[C]{\footnotesize \thepage}      
  \renewcommand{\headrulewidth}{0.4pt}
}

\fancypagestyle{plain}{
  \fancyhf{}
  \fancyhead[R]{\footnotesize Under Review}
  \fancyfoot[C]{\footnotesize \thepage}
  \renewcommand{\headrulewidth}{0.4pt}
}

\AtBeginDocument{\pagestyle{myhdr}}

\title{GenAR: Next-Scale Autoregressive Generation for Spatial Gene Expression Prediction}

\author{
Jiarui Ouyang$^{1,2*}$,Yihui Wang$^{1*}$,Yihang Gao$^{1,2}$, Yingxue Xu$^1$, Shu Yang$^1$ ~Hao Chen$^{1}\textsuperscript{\Letter}$\\ 
$^1$The Hong Kong University of Science and Technology \\
$^2$Shenzhen Loop Area Institute \\ 
\texttt{\{jouyangag, ywangrm, ygaodh, yxueb, syangcw\}@connect.ust.hk} \\ 
\texttt{jhc@cse.ust.hk}
}

\renewcommand\footnotemark{}

\iclrfinalcopy 
\begin{document}

\maketitle

\def\customfootnotetext#1#2{{%
  \let\thefootnote\relax
  \footnotetext[#1]{#2}}}
\customfootnotetext{1}{\textsuperscript{*}Equal contribution.}
\customfootnotetext{2}{\textsuperscript{\Letter}Co-corresponding author.}

\begin{abstract}
Spatial Transcriptomics (ST) offers spatially resolved gene expression but remains costly.
Predicting expression directly from widely available Hematoxylin and Eosin (H\&E) stained images presents a cost-effective alternative.
However, most computational approaches (i) predict each gene independently, overlooking co-expression structure, and (ii) cast the task as continuous regression despite expression being discrete counts. This mismatch can yield biologically implausible outputs and complicate downstream analyses. 
We introduce \textbf{GenAR}, a multi-scale autoregressive framework that refines predictions from coarse to fine. GenAR (a) clusters genes into hierarchical groups to expose cross-gene dependencies, (b) models expression as \emph{codebook-free} discrete token generation to directly predict raw counts, and (c) conditions decoding on fused histological and spatial embeddings. From an information-theoretic view, the discrete formulation avoids log-induced biases and the coarse-to-fine factorization aligns with a principled conditional decomposition. 
Extensive experimental results on four ST datasets across different tissue types demonstrate that GenAR achieves state-of-the-art performance, offering potential implications for precision medicine and cost-effective molecular profiling. Code is publicly available at \href{https://github.com/oyjr/genar}{https://github.com/oyjr/genar}.

\end{abstract}

\section{Introduction}
Spatial Transcriptomics (ST) has emerged as a transformative technology, enabling measurement of gene expression while preserving the spatial organization of cells within tissue samples~\cite{jain2024spatial,Rao_Barkley_França_Yanai_2021,Xiao_Yu_2021}. 
Unlike traditional bulk RNA sequencing, which averages gene expression across entire tissue samples and discards spatial context, ST maintains the spatial relationships among cells and their molecular profiles.
This spatially resolved approach reveals how gene expression patterns vary across different tissue regions, providing molecular insights that complement conventional morphological assessment such as Hematoxylin and Eosin (H\&E) staining~\cite{abmil,mambamil,chen2025stimage,xu2023multimodal}.
The impact of ST technology extends across multiple biomedical domains, such as cancer research, where it identifies spatially distinct tumor subregions~\cite{Bera_Schalper_Rimm_Velcheti_Madabhushi_2019}.

However, ST technology faces significant practical barriers that limit its widespread adoption. Current ST protocols require specialized laboratory equipment, extensive technical expertise, and considerable time investment.
Per-sample costs often range from hundreds to thousands of dollars, making large-scale studies financially challenging~\cite{rao2021exploring}.
These constraints have resulted in relatively small ST datasets, whereas H\&E images are abundant and inexpensive to obtain.
This scarcity of ST data further reduces the practical utility of this technology and hinders comprehensive spatial studies across diverse tissue types and disease conditions.

To address this challenge, several computational methods have been proposed for predicting spatial gene expression directly from histopathological images. 
Early studies such as ST-Net~\cite{he2020integrating} and Hist2ST~\cite{zeng2022spatial} established the basic framework for linking morphological features to molecular profiles. 
Subsequent work includes BLEEP~\cite{xie2023spatially}, which implemented bi-modal embedding with contrastive learning, TRIPLEX~\cite{chung2024accurate}, which integrates multi-resolution feature, and M2OST~\cite{wang2025m2ost} with multimodal and multi-scale strategies. 
More recently, STEM~\cite{zhu2025diffusion} attempted to solve this problem using conditional diffusion models from a generative modeling perspective.

Despite these significant advances, existing methods share several limitations: 
First, many methods independently predict the expression of each gene, underutilizing cross-gene dependencies. 
Genes rarely function in isolation but rather operate in concert through regulatory networks, signaling pathways, and co-expression modules~\cite{barabasi2004network}. 
Treating genes as independent targets can therefore miss biologically meaningful interactions.

Second, existing methods face challenges in maintaining biological interpretability due to their modeling of gene expression as continuous regression tasks.
Gene expression is recorded as nonnegative integer counts that approximate the number of mRNA molecules per spot or cell, typically ranging from zero to several thousand.
These raw counts carry important meaning for biological applications such as differential expression and pathway enrichment~\cite{love2014moderated}. 
However, current approaches apply a $\log$ transformation to gene expression data, converting them into continuous floating-point numbers (typically 0-15) for prediction. 
This transformation departs from the discrete count scale used in biological analyses and may lead to predictions that cannot be directly interpreted in terms of molecule counts.

To address these challenges, we propose GenAR (\textbf{Gen}e expression prediction via next-scale \textbf{A}uto\textbf{R}egressive), a progressive multi-scale autoregressive framework that overcomes the limitations of existing approaches.
Specifically, GenAR tackles the aforementioned problems as follows:
First, rather than predicting genes independently, we cluster genes into coarse-to-fine groups and perform sequential prediction across scales; each scale conditions on all prior predictions to encode cross-gene structure and progressively refine estimates.
Second, our framework directly predicts raw gene expression counts, keeping biological meaning intact and allowing direct use in biological analyses. 
Third, we cast prediction as discrete token generation rather than continuous regression—an information-theoretic, entropy-preserving view that avoids log-induced bias and aligns with a conditional probability decomposition.

Our main contributions can be summarized as follows:
\begin{itemize}
    \item We propose a progressive multi-scale autoregressive framework for spatial gene expression prediction that decomposes the prediction task into sequential scales from coarse to fine granularity.
    
    \item We develop a discrete token generation approach that directly predicts raw gene expression counts through a codebook-free approach, preserving biological interpretability and enabling direct use in downstream analyses.
    
    \item GenAR demonstrates state-of-the-art performance on four spatial transcriptomics datasets, outperforming existing methods across standard evaluation metrics.
\end{itemize}

\section{Related Work}

\subsection{Gene Expression Prediction}


Initial work in this field focused on connecting tissue morphology with molecular profiles.
ST-Net~\cite{he2020integrating} applied the DenseNet architecture to extract features from H\&E stained images for gene expression prediction.
Hist2ST~\cite{zeng2022spatial} advanced this approach by combining convolutional networks, Transformers, and graph neural networks to better capture complex spatial relationships and cellular interactions in tissue samples, demonstrating the importance of modeling spatial context in gene expression prediction.
Histogene~\cite{pang2021leveraging} brought the Vision Transformer architecture to this domain, leveraging self-attention mechanisms to capture long-range dependencies in histopathological images that traditional convolutional approaches might miss.

As the field evolved, specialized methods such as BLEEP~\cite{xie2023spatially} introduced bi-modal embeddings and contrastive learning, aligning histopathological and gene expression data more effectively.
EGN~\cite{yang2023exemplar} utilized exemplar-guided networks for efficient spatial transcriptomics analysis by learning from representative samples, reducing computational overhead while maintaining prediction accuracy. 

Recent work has explored multimodal and multi-scale strategies to further improve performance. 
TRIPLEX~\cite{chung2024accurate} designed a multi-resolution framework with three specialized encoders to capture local patch features, spatial context, and tissue-level patterns at different scales. 
UMPIRE~\cite{han2024towards} employed contrastive learning to align image and gene expression representations using large-scale paired datasets, showing that scaling up training data significantly improves model generalization. 
M2OST~\cite{wang2025m2ost} enhanced prediction accuracy by integrating multimodal information and multi-scale feature representations, STEM~\cite{zhu2025diffusion} attempted to solve this problem from a generative modeling perspective using conditional diffusion models, treating gene expression prediction as a generation task conditioned on histological features. 

\subsection{Next-Scale Autoregressive Generation}

Autoregressive generation has achieved remarkable success in natural language processing and computer vision. 
Early visual autoregressive approaches treated images as sequences of pixels~\cite{van2016conditional}, but suffered from computational inefficiency.
Vector Quantized Variational AutoEncoder (VQ-VAE)~\cite{van2017neural} addressed this by representing images as discrete token sequences through quantization, establishing a two-stage paradigm of discretization followed by autoregressive prediction.

Recently, VAR~\cite{tian2024visual} proposed next-scale prediction, generating images progressively from coarse to fine scales rather than sequential token-by-token generation.
VAR has inspired extensions across diverse applications~\cite{ma2024star,qu2025visual,chen2025visual}, establishing next-scale autoregressive generation as a promising paradigm.
These methods follow a two-stage paradigm where a VQ-VAE first discretizes continuous visual data into codebook tokens, which are then predicted by an autoregressive model. 
This approach is necessitated by the continuous nature of visual data, requiring explicit discretization to enable autoregressive modeling.

However, gene expression prediction presents a unique opportunity to leverage the inherently discrete nature of the data. 
Since gene expression counts are naturally discrete integer values, we can directly apply autoregressive modeling without requiring a separate codebook learning stage. 
This yields a codebook-free end-to-end pipeline that avoids encode–decode reconstruction loss and retains a simpler assumption with fewer moving parts.

\section{Methodology}
\mypara{Problem Formulation.}
Let $\mathcal{G}=\{1,\ldots,n\}$ index the genes. For each spatial location $u$ (spot), we observe an H\&E image patch $I_u \in \mathbb{R}^{\mathcal{H} \times \mathcal{W} \times 3}
$ and its coordinates $S_u \in \mathbb{R}^2$.
The target is a vector of nonnegative integer counts $\mathbf{y}_u \in \mathbb{N}_0^{n}$, where $y_{u,g}$ denotes the expression count of gene $g \in \mathcal{G}$ at location $u$~\cite{anders2010differential}.
Given a dataset $\mathcal{D}=\{(I_u, S_u, \mathbf{y}_u)\}_{u=1}^{N}$, the goal is to learn a mapping that predicts gene expression counts $\hat{\mathbf{y}}_u = \mathrm{GenAR}_{\theta}(I_u, S_u)$. Training minimizes the expected loss over $\mathcal{D}$:
\begin{equation}
\theta^{\star} = \arg\min_{\theta}\;\frac{1}{N}\sum_{u=1}^{N}\,\mathcal{L}(\mathbf{y}_u, \hat{\mathbf{y}}_u)
\end{equation}
where $\mathcal{L}$ is the multi-scale loss function defined in Section~\ref{sec:loss}.

\begin{figure*}[t]
\centering
\includegraphics[width=0.98\textwidth]{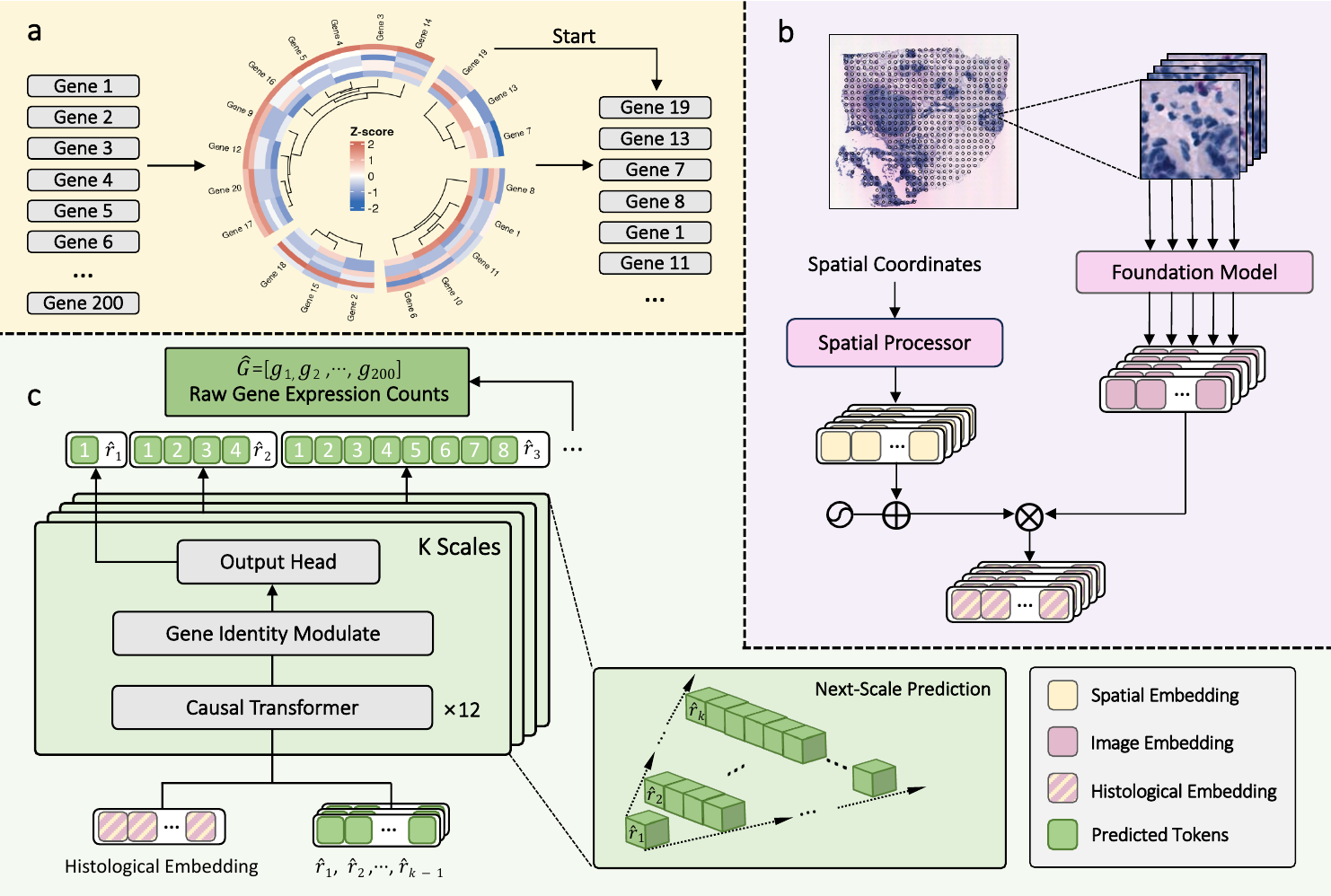}
\caption{Overall architecture of GenAR. 
(a) Genes are clustered into hierarchical groups from coarse to fine granularity. 
(b) Image and spatial features are fused to generate histological embeddings. 
(c) Multi-scale autoregressive generation progressively refines predictions across scales.}
\label{fig:overall}
\end{figure*}

\mypara{Overview of GenAR.} 
An overview of GenAR is shown in Figure~\ref{fig:overall}. 
Genes are reordered into hierarchical clusters based on spatial expression patterns, progressing from major gene groups to smaller nested
subgroups. 
Given an H\&E patch \(I_u\) and its coordinates \(S_u\), we first extract histopathological features using a pre-trained foundation model~\cite{chen2024towards}. 
We then incorporate spatial context by applying a sinusoidal positional encoding to \(S_u\). 
Both modalities are processed through a fusion module to obtain a final histological embedding \(H \in \mathbb{R}^{768}\).

GenAR employs a progressive multi-scale autoregressive framework to capture cross-gene structure that independent regression baselines often underutilize. 
We define $K$ scales with hierarchical gene groups $\{\mathcal{C}^{(1)},\ldots,\mathcal{C}^{(K)}\}$ from coarse to fine, starting from a single global group and refining to smaller groups and finally individual genes.
At each scale $k$, the model predicts grouped expressions $\mathbf{y}^{(k)}_u$ conditioned on all previously generated coarser outputs $\mathbf{y}^{(<k)}_u$.

At each scale, we represent gene expression counts as discrete tokens and map them to dense vectors via a learned embedding layer. 
The resulting sequence is processed by a causal Transformer decoder that is conditioned on the histological embedding \(H\) through adaptive layer normalization (AdaLN)~\cite{dhariwal2021diffusion}. 
We further apply feature-wise linear modulation, where gene-identity embeddings produce scale and shift parameters to inject gene-specific inductive bias into the model. 
Finally, the decoder outputs token logits for the current scale, which are then converted into integer expression counts.

\begin{algorithm}[h!]
\caption{GenAR Training Process}
\label{alg:genar_training_final}
\begin{algorithmic}[1]
\Require Histology patches $I_u$, Spatial coordinates $S_u$, Ground-truth counts $\mathbf{y}$
\Ensure Final training loss $L_{\text{final}}$
\State $H \gets \text{ConditionProcessor}(I_u, S_u)$ \Comment{Fuse multi-modal context}
\State $\mathcal{T} \gets \text{CreateHierarchicalTargets}(\mathbf{y})$
\Comment{Prepare multi-scale ground truth}
\State $\mathcal{E}_{\text{outputs}} \gets \emptyset,\quad L_{\text{total}} \gets 0$
\For{\text{each scale } $k \in \{1, \dots, K\}$ with dimension $d_k$}
    \If{$k=1$}
        \State $X_{\text{context}} \gets \text{[START\_TOKEN]}$
    \Else
        \State $G_{\text{context}} \gets \text{GetTokensFromTargets}(\mathcal{T}_{<k})$ \Comment{Teacher forcing with cumulative history}
        \State $X_{\text{context}} \gets \text{Concat}(\text{[START\_TOKEN]}, \text{GeneEmbed}(G_{\text{context}}))$
    \EndIf
    \State $X_{\text{init}} \gets \text{GeneUpsampling}(\mathcal{E}_{\text{outputs}}, k)$ \Comment{Initialize current scale's targets}
    \State $X \gets \text{Concat}(X_{\text{context}}, X_{\text{init}}) + \text{PosEmbed}(k) + \text{ScaleEmbed}(k)$
    \State $X_{\text{hidden}} \gets \text{Transformer}(X, H, \text{CausalMask})$
    \State $\text{Logits} \gets \text{OutputHead}(\text{FiLM}(\text{SliceLastTokens}(X_{\text{hidden}}, d_k), \text{GeneIdentity}(k)))$
    \If{$k < K$}
        \State $L_k \gets \text{SoftKLLoss}(\text{Logits}, \mathcal{T}_k)$ \Comment{Group-level soft supervision}
    \Else
        \State $L_k \gets \text{GaussianNLL}(\text{CountHead}(\text{Logits}), \mathcal{T}_k;\ \sigma^2=\alpha\,\text{CountHead}(\text{Logits})+\beta)
$ \Comment{Count-level heteroscedastic loss}
    \EndIf
    \State $L_{\text{total}} \gets L_{\text{total}} + L_k$
    \State $\mathcal{E}_{\text{outputs}}.\text{Append}(\text{GeneEmbed}(\text{ArgMax}(\text{Logits})))$ \Comment{Update state for next scale}
\EndFor
\State $L_{\text{final}} \gets L_{\text{total}} / K$
\State \Return $L_{\text{final}}$
\end{algorithmic}
\end{algorithm}

\mypara{Gene Clustering and Histological Embeddings.} 
We cluster genes based on their spatial expression patterns in the training set. 
Using $k$-means on $Z$-score normalized expression profiles, we first group the 200 genes into 4 major clusters, then subdivide each cluster into smaller groups of approximately 12 genes.

The fusion module applies layer normalization to histopathological features $\phi(I_u) \in \mathbb{R}^{1024}$, followed by two linear layers with GELU activation and dropout regularization. 
Spatial information $S_u \in \mathbb{R}^2$ undergoes sinusoidal positional encoding to capture spatial relationships, followed by linear projection and normalization. 
The processed features are concatenated and projected to the final histological embeddings dimension $H \in \mathbb{R}^{768}$.

Gene expression counts are mapped to dense representations through a learned embedding layer $E_{\text{gene}} \in \mathbb{R}^{\text{vocab\_size} \times 768}$. 
Considering that gene expression counts typically range from 0 to several thousand, we adopt a fixed-size vocabulary to cover this range. 
Gene identity embeddings $E_{\text{identity}} \in \mathbb{R}^{n 
\times 768}$ capture the characteristics and functional properties of each gene. 
Gene modulation is achieved through feature-wise linear modulation, where gene identity embeddings are transformed to generate scaling and shift parameters that modulate the hidden representations.

\mypara{Progressive Multi-Scale Generation.} 
The autoregressive generation process can be formalized as:
\begin{equation}
p(\mathbf{y}\mid H) \;=\; \prod_{k=1}^{K} p\big(\mathbf{y}^{(k)} \,\big|\, H,\, \mathbf{y}^{(<k)}\big)
\end{equation}
where $\mathbf{y}^{(k)}$ denotes expressions at scale $k$, $\mathbf{y}^{(<k)}$ denotes all previous-scale outputs, and $H$ represents the histological embeddings derived from histopathological patches and spatial information.

We design $K$ sequential scales to capture gene expression relationships at different granularities, i.e., a structured conditional factorization from global to gene-level interactions.
At each scale $k$, genes are divided into $d_k$ groups, where each group contains consecutive genes from the cluster gene ordering. 
The number of groups increases progressively across scales: the first scale uses a single group representing global transcriptional activity across all genes, intermediate scales gradually increase the number of groups to capture finer-grained patterns, and the final scale contains individual genes for precise prediction. 
We design this hierarchical decomposition to allow GenAR to establish dependencies between genes at different levels of granularity, moving from global transcriptional context to specific gene interactions.

The autoregressive property applies across scales, where predictions at each scale are conditioned on all previously generated coarser-grained information.
At each scale, gene expression values are tokenized and embedded, then processed through a causal Transformer architecture conditioned on the embeddings $H$.
Our approach is codebook-free, directly predicting integer gene expression counts without requiring vector quantization or discrete codebook learning stages. This eliminates potential information loss associated with codebook reconstruction and enables end-to-end training.

\begin{figure*}[t]
\centering
\includegraphics[width=1.0\textwidth]{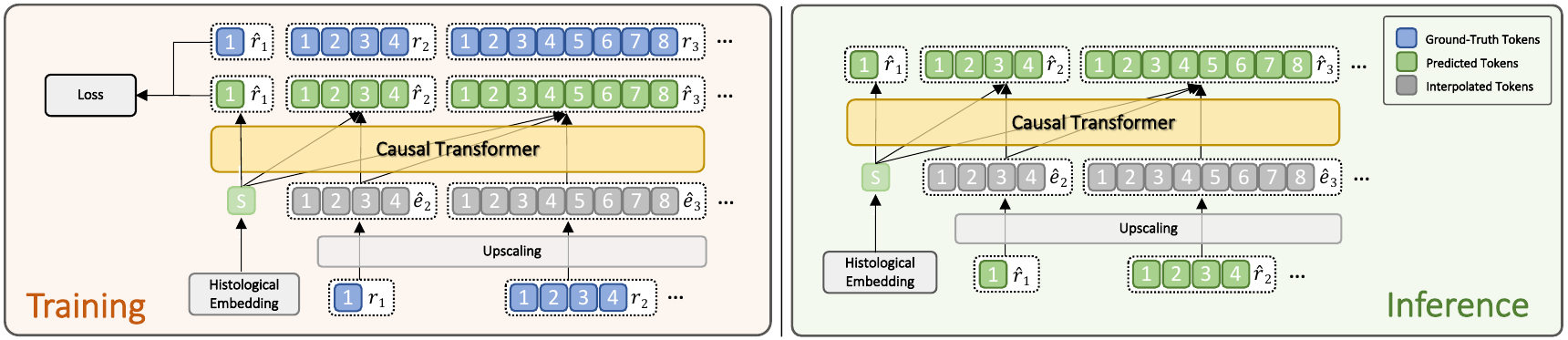}
\caption{Progressive multi-scale generation process, illustrating sequence construction and upsampling initialization during training and inference phases.}
\label{fig:autoregressive}
\end{figure*}

As illustrated in Figure~\ref{fig:autoregressive}, our framework operates differently during training and inference phases. During training (left panel), the model learns to predict tokens at each scale using ground-truth information from previous scales. The process begins with a start token, followed by ground-truth tokens from completed scales, and interpolated tokens that provide initialization for the current scale prediction.

During inference (right panel), the model generates predictions autoregressively across scales. At each scale $k$, the input sequence is constructed as:
\begin{equation}
X_k = [\text{start\_token}, \hat{\mathbf{y}}^{(1)}, \ldots, \hat{\mathbf{y}}^{(k-1)}, \text{interpolated\_tokens}_k]
\end{equation}
where $\hat{\mathbf{y}}^{(j)}$ represents previously generated tokens from scale $j$. The interpolated tokens are obtained through upsampling operations from the previous scale's embeddings, providing contextual initialization for current scale generation.

\mypara{Multi-Scale Loss Function.}
\label{sec:loss}
Under the hierarchical factorization, the negative log-likelihood decomposes across scales:
\begin{equation}
\mathbb{E}_{q}\!\big[-\log p_\theta(\mathbf{y}\mid H)\big]
=\sum_{k=1}^{K}\mathbb{E}_{q}\!\big[-\log p_\theta(\mathbf{y}^{(k)}\mid H,\mathbf{y}^{(<k)})\big]
=\sum_{k=1}^{K}\mathrm{KL}\!\big(q^{(k)}\,\Vert\,p_\theta^{(k)}\big)+\text{const},
\label{eq:ce_kl}
\end{equation}
where $q^{(k)}$ is the target distribution at scale $k$ (group levels use temperature-smoothed targets derived from adaptive pooling; the final level reduces to a sharp target over counts) and $p_\theta^{(k)}$ is the model distribution (softmax over logits).

At intermediate scales, we supervise grouped targets obtained by $\mathbf{y}^{(k)}=\mathrm{AdaptiveAvgPool1d}(\mathbf{y}, d_k)$ for $k<K$
and convert them to soft distributions via temperature smoothing, optimizing $\mathrm{KL}(q^{(k)}\Vert p_\theta^{(k)})$.  
At the final scale, we use a count-level likelihood with expression-dependent variance $\sigma^2=\alpha\mu+\beta$ (equivalently a KL to a Gaussian family up to a constant) to capture heteroscedasticity while preserving count semantics; here $\text{CountHead}(\cdot)$ maps final-scale logits to the mean $\hat{\mu}$ used in the Gaussian NLL.
The overall objective averages losses across scales:
\begin{equation}
\mathcal{L}_{\text{total}}=\frac{1}{K}\sum_{k=1}^{K}\mathcal{L}_k.
\end{equation}

\section{Experiments}

\subsection{Datasets}

We conducted experiments on four different spatial transcriptomics datasets selected from the HEST-1k database~\cite{jaume2024hest}, spanning multiple tissue types and disease states.

\textbf{HER2ST dataset}~\cite{andersson2021spatial} contains breast cancer tissue slides with spatial spots of 100~$\mu$m diameter. 
This dataset consists of multiple pathology images with a total of 13{,}594 spots, each containing gene expression profiles. 
The tissue samples include normal breast tissue and cancerous regions. 
In our experiments, we used the SPA148 slide as the test set, with the remaining slides used for training.

\textbf{Human Prostate Cancer (PRAD) Visium dataset}~\cite{erickson2022spatially} contains 23 prostate cancer tissue slides sequenced using the 10x Genomics Visium platform. 
The spatial spots have a size of 55~$\mu$m, with the number of spots per slide ranging from 1,418 to 4,079. 
We used the MEND145 slide as the test set, with the remaining slides used for training.

\textbf{Kidney Visium dataset}~\cite{lake2023atlas} contains 23 kidney tissue slides from samples representing three pathological states: healthy controls, chronic kidney disease, and acute kidney injury. 
The data was acquired using Visium technology with spatial spots of 55~$\mu$m size, and the number of spots per slide ranges from 315 to 4,159. 
The samples cover both cortical and medullary anatomical regions of the kidney. 
We used the NCBI697 slide as the test set, with the remaining slides used for training.

\textbf{Healthy Mouse Brain dataset}~\cite{vicari2024spatial} contains 14 Visium samples from healthy adult mouse brain tissue. 
Each slide contains 2,675 to 3,617 spatial spots with a spot size of 55~$\mu$m. 
We selected the NCBI667 slide as the test set, with the remaining slides used for training.

\subsection{Data Preprocessing and Evaluation Metrics}
\mypara{Data Preprocessing.} 
For all four datasets, we applied a consistent preprocessing pipeline~\cite{zhu2025diffusion}. 
We selected the top 200 genes from the intersection of highly expressed and highly variable genes for evaluation.
For image processing, we used a patch size of 224 $\times$ 224 pixels for all datasets, with each patch corresponding to one spatial spot. Our model extracts histopathological image features using UNI~\cite{chen2024towards}.
For gene expression count processing, while other baseline models applied $\log_2$ transformation, following~\cite{jaume2024hest}, our model directly predicts raw gene expression counts without $\log_2$ transformation during training. 
To ensure consistent evaluation with other models, we then applied $\log_2$ transformation to our model outputs after prediction for metric calculation.

\mypara{Evaluation Metrics.} 
We used multiple evaluation metrics to assess model prediction performance.
We used PCC-10, PCC-50, and PCC-200 metrics, representing the average PCC values of the top 10, 50, and 200 genes with the highest Pearson correlation coefficients in the prediction results, respectively. 
For a single gene $g$, the PCC is calculated as:

\begin{equation}
\operatorname{PCC}_g = \frac{\operatorname{Cov}(Y_g, \hat{Y}_g)}{\sqrt{\operatorname{Var}(Y_g) \cdot \operatorname{Var}(\hat{Y}_g)}}
\end{equation}

where $Y_g$ and $\hat{Y}_g$ represent the true and predicted expression values for gene $g$, respectively, $\operatorname{Cov}(\cdot)$ denotes covariance, and $\operatorname{Var}(\cdot)$ denotes variance.

Moreover, we calculated Mean Squared Error (MSE) and Mean Absolute Error (MAE) to evaluate the overall prediction accuracy of the model. 
MSE computes the average squared error between predicted and true values across all spatial spots and genes. 
MAE computes the average absolute error between predicted and true values across all spatial spots and genes.

\begin{table*}[t]
\centering
\scriptsize
\setlength{\tabcolsep}{3pt}
\begin{tabular}{l|ccccc|ccccc}
\toprule
\multirow{2}{*}{Method} & \multicolumn{5}{c|}{HER2ST} & \multicolumn{5}{c}{PRAD} \\
\cmidrule(lr){2-6} \cmidrule(lr){7-11}
& PCC-10↑ & PCC-50↑ & PCC-200↑ & MSE↓ & MAE↓ & PCC-10↑ & PCC-50↑ & PCC-200↑ & MSE↓ & MAE↓ \\
\midrule
BLEEP~\cite{xie2023spatially} & 0.773 & 0.714 & 0.565 & 1.243 & 0.833 & 0.580 & 0.510 & 0.316 & 2.475 & 1.091 \\
M2OST~\cite{wang2025m2ost} & 0.810 & 0.759 & 0.660 & 1.151 & 0.820 & 0.602 & 0.551 & 0.442 & 1.290 & 0.862 \\
TRIPLEX~\cite{chung2024accurate} & 0.783 & 0.714 & 0.586 & 1.212 & 0.857 & 0.620 & 0.544 & 0.423 & 1.319 & 0.836 \\
STEM~\cite{zhu2025diffusion} & 0.831 & 0.770 & 0.625 & 1.199 & 0.787 & 0.636 & 0.555 & 0.403 & 1.457 & 0.857 \\
\textbf{GenAR(Ours)} & \textbf{0.842} & \textbf{0.784} & \textbf{0.663} & \textbf{1.082} & \textbf{0.745} & \textbf{0.702} & \textbf{0.650} & \textbf{0.512} & \textbf{1.191} & \textbf{0.771} \\
\bottomrule
\end{tabular}
\caption{Experimental results on HER2ST and PRAD datasets. The best results are highlighted in \textbf{bold}. ↑ indicates higher is better, ↓ indicates lower is better.}
\label{tab:results_1}
\end{table*}

\begin{table*}[t]
\centering
\scriptsize
\setlength{\tabcolsep}{3pt}
\begin{tabular}{l|ccccc|ccccc}
\toprule
\multirow{2}{*}{Method} & \multicolumn{5}{c|}{Kidney} & \multicolumn{5}{c}{Healthy Mouse Brain} \\
\cmidrule(lr){2-6} \cmidrule(lr){7-11}
& PCC-10↑ & PCC-50↑ & PCC-200↑ & MSE↓ & MAE↓ & PCC-10↑ & PCC-50↑ & PCC-200↑ & MSE↓ & MAE↓ \\
\midrule
BLEEP~\cite{xie2023spatially} & 0.500 & 0.422 & 0.314 & 1.926 & 0.945 & 0.342 & 0.280 & 0.156 & 1.591 & 0.987 \\
M2OST~\cite{wang2025m2ost} & 0.494 & 0.447 & 0.318 & 1.785 & 0.925 & 0.456 & 0.387 & 0.231 & 1.148 & 0.861 \\
TRIPLEX~\cite{chung2024accurate} & 0.542 & 0.469 & 0.336 & 1.732 & 0.887 & 0.501 & 0.445 & 0.312 & 1.157 & 0.822 \\
STEM~\cite{zhu2025diffusion} & 0.567 & 0.483 & 0.322 & 1.832 & 0.997 & 0.526 & 0.452 & 0.331 & 1.235 & 0.864 \\
\textbf{GenAR(Ours)} & \textbf{0.589} & \textbf{0.514} & \textbf{0.354} & \textbf{1.636} & \textbf{0.871} & \textbf{0.568} & \textbf{0.503} & \textbf{0.367} & \textbf{1.138} & \textbf{0.805} \\
\bottomrule
\end{tabular}
\caption{Experimental results on Kidney and Mouse Brain datasets. The best results are highlighted in \textbf{bold}. ↑ indicates higher is better, ↓ indicates lower is better.}
\label{tab:results_2}
\end{table*}

\mypara{Implementation Details.} 
Our models are trained with the Adam optimizer~\cite{kingma2014adam} with a learning rate of 1e-4 and a batch size of 64.
For these four datasets, we configured 6 hierarchical scale dimensions (1, 4, 8, 40, 100, 200) for multi-scale feature extraction, targeting the prediction of 200 gene expression counts. 
The main hyperparameters of our model include model depth, model width, and the number of heads in self-attention mechanisms. 
Increased model complexity is often accompanied by higher computational resource requirements, our model design balances prediction performance while considering computational efficiency.
Experiments were conducted in PyTorch on NVIDIA H100 (80\,GB) GPUs. Baselines used the same preprocessing and tuning protocol. 
Results were obtained with fixed seeds. 
Code and configuration files will be released.

\subsection{Experimental Results}
We evaluated the performance of our proposed method on four different spatial transcriptomics datasets and compared it with multiple baseline methods. The experimental results are shown in Table~\ref{tab:results_1} and Table~\ref{tab:results_2}. Our method achieves the best performance across all datasets.

As shown in Table~\ref{tab:results_1}, on the HER2ST dataset containing breast cancer tissue slides with 100~$\mu$m spatial spots, our method achieves PCC-10, PCC-50, and PCC-200 scores of 0.842, 0.784, and 0.663, outperforming the best baseline method STEM by 1.3\%, 1.8\%, and 6.1\%, respectively. 
The MSE and MAE are reduced by 9.8\% and 5.3\% compared to STEM.
On the PRAD dataset containing prostate cancer tissue slides with 55~$\mu$m spatial spots, our method demonstrates more significant improvements, achieving PCC-10, PCC-50, and PCC-200 scores of 0.702, 0.650, and 0.512, surpassing STEM by 10.4\%, 17.1\%, and 27.0\%, respectively. 
The MSE and MAE reductions are 18.3\% and 10.0\%, respectively.

As shown in Table~\ref{tab:results_2}, on the Kidney dataset covering three pathological states with 55~$\mu$m spatial spots, our method achieves PCC-10, PCC-50, and PCC-200 scores of 0.589, 0.514, and 0.354, outperforming STEM by 3.9\%, 6.4\%, and 9.9\%, respectively. 
The MSE and MAE are reduced by 10.7\% and 12.6\%.
On the Mouse Brain dataset containing healthy adult mouse brain tissue with 55~$\mu$m spatial spots, our method achieves PCC-10, PCC-50, and PCC-200 scores of 0.568, 0.503, and 0.367, surpassing STEM by 8.0\%, 11.3\%, and 10.9\%, respectively. The MSE and MAE reductions are 7.9\% and 6.8\%.

Cross-dataset analysis shows consistent gains, with larger margins on cancer tissues, indicating that the proposed coarse-to-fine discrete autoregressive formulation is robust across tissue types.




\subsection{Ablation Study}

\mypara{Component Ablation.} 
We perform internal ablation experiments on the PRAD dataset, which contains numerous spatial spots and exhibits high complexity. 
As shown in Table~\ref{tab:ablation_internal}, we separately remove the progressive multi-scale generation framework, gene identity embeddings, and replace the loss function to analyze the contribution of each component.

The experimental results demonstrate that removing the progressive multi-scale generation framework has the most significant impact on model performance, with PCC-10 decreasing from 0.702 to 0.651 and MSE increasing from 1.171 to 1.406. 
This indicates that the multi-scale autoregressive generation process is crucial for capturing gene expression relationships at different granularities. 
After removing gene identity embeddings, the PCC-200 metric decreases from 0.532 to 0.481, demonstrating the importance of gene-specific representations for precise prediction. 
Using cross-entropy loss instead of our designed adaptive Gaussian KL loss and soft-label KL divergence loss results in PCC-10 decreasing from 0.702 to 0.662.
We also observe weaker performance in extremely sparse regions (token rarity/gradient sparsity), motivating pathway/ontology-informed grouping without altering the core framework.

\begin{table}[t]
\centering
\footnotesize
\setlength{\tabcolsep}{1pt}
\begin{tabular}{l|ccccc}
\toprule
Method & PCC-10↑ & PCC-50↑ & PCC-200↑ & MSE↓ & MAE↓ \\
\midrule
w/o Multi-scale & 0.651 & 0.601 & 0.493 & 1.406 & 0.779 \\
w/o Gene identity & 0.683 & 0.625 & 0.481 & 1.281 & 0.828 \\
w/ Cross-entropy & 0.662 & 0.612 & 0.482 & 1.325 & 0.781 \\
GenAR & \textbf{0.702} & \textbf{0.650} & \textbf{0.512} & \textbf{1.191} & \textbf{0.771} \\
\bottomrule
\end{tabular}
\caption{Ablation study results on the PRAD dataset. The best results are highlighted in \textbf{bold}. ↑ indicates higher is better, ↓ indicates lower is better.}
\label{tab:ablation_internal}
\end{table}

\begin{figure*}[t]
\centering
\includegraphics[width=1\textwidth]{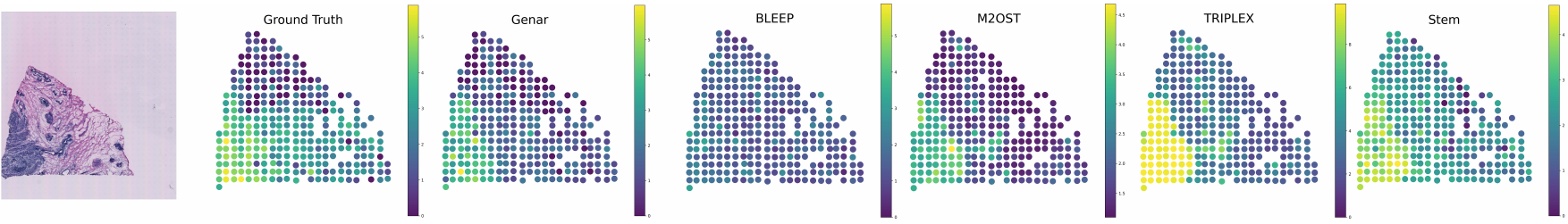}
\caption{Spatial visualization of SSR4 gene expression prediction on HER2ST SPA148 sample. From left to right: histopathological image, ground truth, and predictions from GenAR, BLEEP, M2OST, TRIPLEX, and STEM. Color scale: low (purple/blue) to high (yellow/green) expression.}
\label{fig:vis}
\end{figure*}

\begin{table}[t]
\centering
\footnotesize
\setlength{\tabcolsep}{1pt}
\begin{tabular}{l|ccccc}
\toprule
Method & PCC-10↑ & PCC-50↑ & PCC-200↑ & MSE↓ & MAE↓ \\
\midrule
ResNet-18~\cite{he2016deep} & 0.431 & 0.381 & 0.273 & 1.876 & 0.905 \\
CONCH~\cite{huang2023visual} & 0.451 & 0.391 & 0.286 & 2.110 & 0.949 \\
UNI~\cite{chen2024towards} & 0.597 & 0.526 & 0.383 & 1.741 & 0.871 \\
\textbf{GenAR (Ours)} & \textbf{0.702} & \textbf{0.650} & \textbf{0.512} & \textbf{1.191} & \textbf{0.771} \\
\bottomrule
\end{tabular}
\caption{Performance comparison on raw gene expression count prediction on the PRAD dataset. The best results are highlighted in \textbf{bold}. ↑ indicates higher is better, ↓ indicates lower is better.}
\label{tab:prad_comparison}
\end{table}

\mypara{Foundation Model Ablation.} 
We design a raw gene expression count prediction task and evaluate it across three foundation models and GenAR. 
We compare ResNet-18~\cite{he2016deep}, CONCH~\cite{huang2023visual}, and UNI~\cite{chen2024towards}. 
Our full GenAR model selects the UNI~\cite{chen2024towards} to extract histological features.
For fair comparison, all baselines use identical architectures: input projection to hidden space, two residual blocks with GELU activation, and output heads for discrete gene expression prediction.
Table~\ref{tab:prad_comparison} reports results on the PRAD dataset. 
All three foundation models achieve reasonable performance on this discrete prediction task, while GenAR substantially outperforms them with 33.7\% improvement in PCC-200 over the best baseline, demonstrating the effectiveness of our framework design.

\subsection{Visualization Analysis}
Figure~\ref{fig:vis} shows visualization results on the HER2ST dataset using sample SPA148 for gene SSR4 (Signal Sequence Receptor Subunit 4), which encodes an ER membrane receptor associated with cancer progression. The ground truth exhibits distinct spatial heterogeneity with concentrated high-expression regions (yellow-green) and clearly demarcated low-expression areas (purple-blue).

BLEEP and M2OST generate smooth expression maps with limited dynamic range, failing to capture sharp spatial transitions. 
TRIPLEX and STEM show improved pattern recognition, with STEM preserving better boundaries, though both exhibit oversmoothing in high-expression regions. 
GenAR produces expression predictions that most closely match ground truth, accurately capturing both high-expression zone localization and expression level transitions.

\section{Conclusion}
We propose GenAR, a multi-scale autoregressive framework that reframes spatial gene expression prediction as discrete token generation. 
The discrete formulation preserves biological interpretability and avoids biases of continuous surrogates, while the coarse-to-fine factorization encodes hierarchical dependencies. 
Empirically, GenAR achieves state-of-the-art performance across four datasets. We also note relatively weaker performance in extremely sparse regions, which motivates pathway and ontology informed grouping.
The design is modality agnostic and may extend to proteomics, metabolomics, or other spatiotemporal settings. 
Overall, this compact, codebook-free recipe may inform broader multimodal learning research. Future work will explore the integration of more sophisticated biological priors, such as gene regulatory networks and pathway-level interactions, to further enhance prediction accuracy.



\section*{Acknowledgments}
This work was supported by the National Natural Science Foundation of China (No. 62202403), Hong Kong Innovation and Technology Commission (Project No. MHP/002/22 and ITCPD/17-9), and Research Grants Council of the Hong Kong Special Administrative Region, China (Project No. R6003-22 and C4024-22GF).
\section*{Ethics Statement}
This work uses only publicly available, de-identified spatial transcriptomics datasets and H\&E images; no new human or animal data were collected. We complied with dataset licenses and standard citation practices. The models are released for research use only and are not intended for clinical decision-making. We do not release any sensitive metadata.

\section*{Reproducibility Statement}
We took several steps to support reproducibility. The paper details the method in Section~\ref{sec:loss} (architecture, training objective, multi-scale design), the datasets, gene selection, preprocessing pipeline, and metrics in Section~4.2, and training and implementation details (optimizers, batch sizes, scales, seeds, hardware) in Section~4.3. The Appendix further lists all selected genes for each dataset, ablation settings, and exact hyperparameters and training setup (``Implementation Details and Reproducibility''). Code is publicly available at \href{https://github.com/oyjr/genar}{https://github.com/oyjr/genar}.

\bibliography{iclr2026_conference}
\bibliographystyle{iclr2026_conference}

\newpage
\appendix

\section{Additional Experiment}
\subsection{ccRCC Dataset Results}
We conducted additional experiments on the ccRCC (clear cell Renal Cell Carcinoma) dataset~\cite{meylan2022tertiary} to further validate the generalisation capability of our proposed GenAR framework across different cancer types.

\textbf{ccRCC dataset} contains 24 clear cell renal cell carcinoma tissue slides sequenced using the 10x Genomics Visium platform. Each spatial spot covers an area of 112×112~$\mu$m, and the number of spots per slide varies across samples. The samples are labeled from INT1 to INT24, representing different tissue sections from kidney cancer patients. We used the INT2 slide as the test set, with the remaining slides used for training.

\begin{table}[h]
\centering
\footnotesize
\setlength{\tabcolsep}{2pt}
\begin{tabular}{l|ccccc}
\toprule
Method & PCC-10↑ & PCC-50↑ & PCC-200↑ & MSE↓ & MAE↓ \\
\midrule
BLEEP & 0.366 & 0.288 & 0.202 & 1.853 & 1.238 \\
M2OST & 0.408 & 0.309 & 0.195 & 1.798 & 1.068 \\
TRIPLEX & 0.429 & 0.336 & 0.246 & 1.553 & 0.938 \\
STEM & 0.429 & 0.377 & 0.256 & \textbf{1.422} & 0.932 \\
\textbf{GenAR(Ours)} & \textbf{0.457} & \textbf{0.394} & \textbf{0.276} & 1.465 & \textbf{0.896} \\
\bottomrule
\end{tabular}
\caption{Experimental results on ccRCC dataset. The best results are highlighted in \textbf{bold}. ↑ indicates higher is better, ↓ indicates lower is better.}
\label{tab:ccrcc_results}
\end{table}

\begin{table}[t]
\centering
\footnotesize
\setlength{\tabcolsep}{1pt}
\begin{tabular}{l|ccccc}
\toprule
Scale Design & PCC-10↑ & PCC-50↑ & PCC-200↑ & MSE↓ & MAE↓ \\
\midrule
(200) & 0.622 & 0.561 & 0.445 & 1.446 & 0.797 \\
(1, 20, 200) & 0.681 & 0.638 & \textbf{0.534} & 1.203 & 0.775 \\
(1, 40, 100, 200) & 0.683 & 0.643 & 0.522 & 1.213 & \textbf{0.763} \\
\textbf{(1, 4, 8, 40, 100, 200) (Ours)} & \textbf{0.702} & \textbf{0.650} & 0.512 & \textbf{1.191} & 0.771 \\
\bottomrule
\end{tabular}
\caption{Ablation study on different scale designs for gene expression count prediction on the PRAD dataset. The best results are highlighted in \textbf{bold}. ↑ indicates higher is better, ↓ indicates lower is better.}
\label{tab:scale_ablation}
\end{table}
The results on the ccRCC dataset demonstrate consistent performance improvements, with PCC-10, PCC-50, and PCC-200 scores of 0.457, 0.394, and 0.276, respectively, outperforming the best baseline methods. Our method achieves a 6.5\% improvement in PCC-10 compared to both TRIPLEX and STEM, and shows a 4.5\% improvement in PCC-50 over STEM. For PCC-200, GenAR achieves a 7.8\% improvement compared to STEM. While the MSE is slightly higher than STEM, the MAE is reduced by 3.9\%, indicating better overall prediction accuracy. These results further validate the effectiveness of our progressive multi-scale autoregressive approach across different cancer types.

\subsection{Scale Designs Ablation}

To investigate the impact of different scale designs on prediction performance, we conducted ablation studies on the PRAD dataset with varying hierarchical decompositions. As shown in Table~\ref{tab:scale_ablation}, we compare four different scale configurations: (1) single scale with all 200 genes, (2) three scales with 1, 20, and 200 groups, (3) four scales with 1, 40, 100, and 200 groups, and (4) our proposed six-scale design with 1, 4, 8, 40, 100, and 200 groups.

The results demonstrate that increasing the number of scales generally improves performance. 
The single-scale baseline (200) achieves the lowest performance across most metrics, confirming the importance of hierarchical decomposition. 
The three-scale design shows significant improvements, particularly in PCC-200 (0.534). 
Adding a fourth scale further enhances PCC-10 and PCC-50 performance while achieving the best MAE (0.763). 
Our six-scale design achieves the best overall performance with PCC-10 of 0.702 and PCC-50 of 0.650, representing the optimal balance between hierarchical granularity and model complexity. 
The progressive refinement from global transcriptional context (1 group) to individual gene predictions (200 groups) enables the model to capture dependencies at multiple levels of granularity.



\section{Inference Process}
During inference, GenAR generates predictions autoregressively across scales using previously generated outputs as context (no teacher forcing). At each scale $k$, the input sequence concatenates the start token, embeddings of all past predictions $\hat{\mathbf{y}}^{(1)},\ldots,\hat{\mathbf{y}}^{(k-1)}$, and interpolated tokens that initialize the current scale. The Transformer decoder processes this sequence under a causal mask, produces logits, and the current tokens $\hat{\mathbf{y}}^{(k)}$ are obtained from the predicted distribution (default: greedy $\arg\max$). These tokens are appended to the history to condition subsequent scales, enabling progressive refinement from coarse- to fine-grained predictions.

\begin{algorithm}[h!]
\caption{GenAR Inference Process}
\label{alg:genar_inference}
\begin{algorithmic}[1]
\Require Histology patches $I_u$, Spatial coordinates $S_u$
\Ensure Final prediction $\hat{\mathbf{y}}^{(K)}$
\State $H \gets \text{ConditionProcessor}(I_u, S_u)$ \Comment{Fuse multi-modal context}
\State $\mathcal{E}_{\text{outputs}} \gets \emptyset$,\quad $\hat{\mathbf{y}}^{(<1)} \gets \emptyset$
\For{\text{each scale } $k \in \{1, \dots, K\}$ with dimension $d_k$}
    \If{$k=1$}
        \State $X_{\text{context}} \gets \text{[START\_TOKEN]}$
    \Else
        \State $X_{\text{context}} \gets \text{Concat}(\text{[START\_TOKEN]}, \text{GeneEmbed}(\hat{\mathbf{y}}^{(<k)}))$
    \EndIf
    \State $X_{\text{init}} \gets \text{GeneUpsampling}(\mathcal{E}_{\text{outputs}}, k)$ \Comment{Initialize current scale}
    \State $X \gets \text{Concat}(X_{\text{context}}, X_{\text{init}}) + \text{PosEmbed}(k) + \text{ScaleEmbed}(k)$
    \State $X_{\text{hidden}} \gets \text{Transformer}(X, H, \text{CausalMask})$
    \State $\text{Logits} \gets \text{OutputHead}(\text{FiLM}(\text{SliceLastTokens}(X_{\text{hidden}}, d_k), \text{GeneIdentity}(k)))$
    \State $\hat{\mathbf{y}}^{(k)} \gets \arg\max(\text{Logits})$ \Comment{Greedy decode; sampling with temperature is optional}
    \State $\hat{\mathbf{y}}^{(<k+1)} \gets \text{Concat}(\hat{\mathbf{y}}^{(<k)}, \hat{\mathbf{y}}^{(k)})$
    \State $\mathcal{E}_{\text{outputs}}.\text{Append}(\text{GeneEmbed}(\hat{\mathbf{y}}^{(k)}))$ \Comment{State for next scale}
\EndFor
\State \Return $\hat{\mathbf{y}}^{(K)}$
\end{algorithmic}
\end{algorithm}

\section{Selected Genes for All Datasets}

We provide the complete selected genes for each of the five datasets used in our experiments. 
The genes were selected based on the intersection of highly expressed and highly variant genes following the preprocessing pipeline described in Section 4.2 Data Preprocessing and Evaluation Metrics.

\subsection{HER2ST Dataset Selected Genes}

The following genes were selected for the HER2ST (breast cancer) dataset:

\begin{footnotesize}
\begin{flushleft}
A2M, ACTB, ACTG1, ACTN4, ADAM15, AEBP1, AES, ALDOA, AP000769.1, APOC1, APOE, ARHGDIA, ATG10, ATP5B, ATP5E, ATP6V0B, AZGP1, B2M, BEST1, BGN, BSG, BST2, C12orf57, C1QA, C3, CALM2, CALML5, CALR, CCT3, CD24, CD63, CD74, CFL1, CHCHD2, CHPF, CIB1, CLDN3, CLDN4, COL18A1, COL1A1, COL1A2, COL3A1, COL6A2, COMP, COPE, COPS9, COX4I1, COX5B, COX6B1, COX6C, COX7C, CRIP2, CST3, CTSB, CTSD, CTTN, CYBA, DBI, DDIT4, DDX5, DHCR24, EDF1, EEF1D, EEF2, EIF4G1, ELOVL1, ENO1, ERBB2, ERGIC1, FASN, FAU, FLNA, FN1, FNBP1L, FTH1, FTL, GAPDH, GNAS, GPX4, GRB7, GRINA, GUK1, H2AFJ, HLA-A, HLA-B, HLA-C, HLA-DRA, HLA-E, HNRNPA2B1, HSP90AA1, HSP90AB1, HSP90B1, HSPA8, HSPB1, IDH2, IFI27, IGFBP2, IGHA1, IGHG1, IGHG3, IGHG4, IGHM, IGKC, IGLC2, IGLC3, INTS1, ISG15, JTB, KDELR1, KRT18, KRT19, KRT7, KRT81, LAPTM4A, LAPTM5, LASP1, LGALS1, LGALS3, LGALS3BP, LMAN2, LMNA, LUM, LY6E, MAPKAPK2, MDK, MGP, MIDN, MIEN1, MLLT6, MMACHC, MMP14, MUC1, MUCL1, MYL6, MYL9, MZT2B, NACA, NBL1, NDUFB9, NUCKS1, NUPR1, ORMDL3, P4HB, PCGF2, PEBP1, PERP, PFDN5, PFKL, PGAP3, PHB, PIP4K2B, PLD3, POSTN, PPDPF, PPP1CA, PPP1R1B, PRDX1, PRRC2A, PRSS8, PSMB3, PSMB4, PSMD3, PTMA, PTMS, PTPRF, RACK1, S100A14, S100A6, S100A8, S100A9, SCAND1, SCD, SDC1, SEC61A1, SEPW1, SERF2, SF3B5, SH3BGRL3, SLC2A4RG, SLC9A3R1, SNRPB, SPARC, SPDEF, SPINT2, SSR2, SSR4, STARD10, STARD3, SUPT6H, SYNGR2, TAGLN, TAPBP, TFF3, TIMP1, TMED9, TMSB10, TPT1, TSPO, TUBB, TXNIP, TYMP, UBA52, UBC, UBE2M, UBL5, UQCRQ, VIM, ZYX.
\end{flushleft}
\end{footnotesize}

We clustered the 200 genes into 32 groups. The group membership is listed below.

\begin{longtable}{p{0.06\linewidth} p{0.9\linewidth}}
\toprule
\textbf{Group} & \textbf{Members} \\
\midrule
0 & ACTB, AES, BST2, CALM2, CALML5, CALR, CIB1, COX6C, CST3, IGFBP2, KDELR1, TUBB \\
1 & ATP5E, IGHA1, IGHG4 \\
2 & IGHG3 \\
3 & GRB7 \\
4 & MLLT6 \\
5 & SNRPB \\
6 & ADAM15, AEBP1, APOC1, ARHGDIA, COL18A1, COL1A1, COL1A2, COL3A1, COPS9, FLNA, FN1, PTPRF \\
7 & C1QA, COX4I1, DDX5, ENO1, FAU, FNBP1L, FTH1, KRT19, KRT7, LGALS3, MIEN1, SSR2 \\
8 & ACTN4, ELOVL1, ERBB2, FASN, HLA\textendash B, HLA\textendash C, HLA\textendash DRA, LGALS3BP, PEBP1, PPDPF, PTMS, RACK1 \\
9 & ERGIC1, FTL, KRT81, LGALS1, LMAN2, MIDN, PERP, PFDN5, PFKL, PTMA, S100A14, SPDEF \\
10 & ATP6V0B, AZGP1, COX5B, CTTN, CYBA, DHCR24, IGKC, IGLC2, PSMD3, S100A6, SSR4, STARD10 \\
11 & DDIT4, HLA\textendash A, HLA\textendash E, IGLC3, KRT18, PHB, PIP4K2B, PPP1R1B, PRDX1, S100A8, SLC2A4RG, SUPT6H \\
12 & ATG10, EEF2, EIF4G1, GAPDH, HSPA8, MZT2B, P4HB, POSTN, PRSS8, PSMB3, SCAND1, TFF3 \\
13 & CD74, DBI, EDF1, HNRNPA2B1, HSP90AB1, HSP90B1, LAPTM5, PGAP3, PSMB4, SEC61A1, SLC9A3R1, STARD3 \\
14 & CLDN4, GNAS, GRINA, LASP1, MMACHC, PCGF2, PPP1CA, PRRC2A, SF3B5, SH3BGRL3, SPARC, SPINT2 \\
15 & B2M, BEST1, CD24, CLDN3, CTSD, GPX4, GUK1, H2AFJ, MGP, MMP14, MYL9, SEPW1 \\
16 & A2M, AP000769.1, CHPF, EEF1D, IFI27, LMNA, LUM, MUC1, MYL6, PLD3, SCD, TAGLN \\
17 & HSP90AA1, IGHM, INTS1, LAPTM4A, MUCL1, NUPR1, ORMDL3, S100A9, SERF2, SYNGR2, TAPBP, TMED9 \\
18 & C3, CHCHD2, COX6B1, CRIP2, JTB, LY6E, MAPKAPK2, MDK, NACA, NDUFB9, SDC1, TXNIP \\
19 & ACTG1, BGN, BSG, CFL1, COPE, COX7C, CTSB, HSPB1, NBL1, TIMP1, TMSB10 \\
20 & COMP \\
21 & IDH2, ISG15 \\
22 & CCT3 \\
23 & COL6A2 \\
24 & ATP5B \\
25 & TSPO \\
26 & CD63 \\
27 & APOE \\
28 & NUCKS1 \\
29 & TPT1 \\
30 & C12orf57 \\
31 & IGHG1 \\
\bottomrule
\end{longtable}

\paragraph{Examples of biologically coherent modules.}
Several groups align with well-known breast cancer and microenvironment programs:
\begin{itemize}
  \item \textbf{HER2/17q12 amplicon}: groups \textbf{8}, \textbf{3}, \textbf{13}, \textbf{7} contain \textit{ERBB2}, \textit{GRB7}, \textit{STARD3}, \textit{PGAP3}, \textit{MIEN1}, which are frequently co\textendash amplified and co\textendash expressed in HER2\textsuperscript{+} tumors~\cite{hongisto2014her2amplicon,kwon2017erbb2coamplicon}.
  \item \textbf{Luminal epithelial/secretory features}: groups \textbf{12}, \textbf{14}, \textbf{15}, \textbf{16}, \textbf{9} include \textit{MUC1}, \textit{TFF3}, \textit{SPDEF}, \textit{CLDN3}/\textit{CLDN4}, \textit{KRT7}/\textit{KRT19}, consistent with luminal programs~\cite{gray2022humanbreast}.
  \item \textbf{Antigen presentation and interferon\textendash stimulated genes}: groups \textbf{8}, \textbf{11}, \textbf{21}, \textbf{18} include \textit{HLA\textendash A/B/C/E/DRA}, \textit{ISG15}, \textit{IFI27}, \textit{LY6E}, reflecting MHC and IFN response modules with prognostic links in breast cancer~\cite{kariri2020isg15,bektas2008isg15}.
  \item \textbf{C1Q\textsuperscript{+} macrophages}: group \textbf{7} contains \textit{C1QA} together with \textit{LGALS3}, consistent with C1Q\textsuperscript{+} TAM subsets described in breast tumors~\cite{zhang2024tam}.
  \item \textbf{CAF/ECM remodeling and smooth\textendash muscle}: groups \textbf{6}, \textbf{12}, \textbf{14}, \textbf{19}, \textbf{23}, \textbf{16} include \textit{COL1A1/1A2/3A1/6A2}, \textit{FN1}, \textit{POSTN}, \textit{SPARC}, \textit{TAGLN}, \textit{LUM}, typical of stromal and myofibroblast programs~\cite{chen2021caf,cords2023caf}.
  \item \textbf{Plasma\textendash cell immunoglobulins}: groups \textbf{1}, \textbf{2}, \textbf{10}, \textbf{11}, \textbf{31} contain \textit{IGH*} and \textit{IGK/IGL} genes, consistent with plasma\textendash cell infiltration and known prognostic associations~\cite{whiteside2012igkc,yeong2018plasmacells}.
\end{itemize}

\subsection{Kidney Dataset Selected Genes}

The following genes were selected for the Kidney dataset:

\begin{footnotesize}
\begin{flushleft}
A1BG, A2M, ACADVL, ACTA2, ACTB, ACTG1, ADGRG1, ADIRF, AEBP1, ALDOB, ANPEP, ANXA2, APOE, APP, AQP1, AQP2, ASS1, ATP1A1, ATP1B1, ATP5F1D, ATP5MC3, ATP5MD, ATP5ME, ATP5MF, ATP5MPL, ATP6V0C, B2M, BCAM, BGN, BSG, C7, CA2, CALB1, CALM2, CANX, CD151, CD24, CD74, CD81, CFL1, CHCHD10, CIRBP, CKB, CLCNKB, CLU, COL1A2, COL3A1, COL4A2, COX5A, COX5B, COX6B1, COX6C, COX7A2, COX7B, COX7C, CRIM1, CRYAB, CST3, CTSB, CTSH, CXCL14, CYSTM1, DCN, DDX5, DEFB1, DSTN, DUSP1, DYNLL1, EEF1D, EEF1G, EEF2, EIF3K, ENG, EPAS1, EZR, FLNA, FTH1, FTL, FXYD2, GABARAP, GATM, GPX3, GSTP1, H3F3A, HINT1, HLA-A, HLA-B, HLA-C, HLA-DRA, HLA-DRB1, HLA-E, HNRNPA1, HNRNPA2B1, HSD11B2, HSPA8, HSPB1, HTRA1, IDH2, IFITM2, IFITM3, IGFBP4, IGFBP5, IGFBP7, IGHA1, IGHG1, IGHG3, IGHG4, IGKC, IGLC1, IGLC2, IGLC3, ITM2B, KNG1, LAMP1, LAMTOR5, LAPTM4A, LDHA, LGALS1, LRP2, LUM, MAL, MALAT1, MGP, MGST3, MIOX, MMP7, MUC1, MYL6, MYL9, NAT8, NDRG1, NDUFA1, NDUFA13, NDUFA4, NDUFB2, NDUFB7, NDUFB8, NDUFB9, NEAT1, NME2, OAZ1, OGDHL, OST4, P4HB, PCK1, PDZK1IP1, PEBP1, PEPD, PFN1, PGK1, PIGR, PODXL, PPP1R1A, PTGDS, PTH1R, REN, RHOA, RNASE1, RTN4, S100A10, S100A2, S100A6, SAT1, SELENOP, SERPINA1, SERPINA5, SFRP1, SLC12A1, SLC12A3, SLC13A3, SLC25A3, SLC25A5, SLC25A6, SLC3A1, SLC5A12, SOD1, SOD2, SPARC, SPINK1, SPP1, SRP14, SSR4, SUCLG1, TAGLN, TIMP1, TIMP3, TMA7, TMSB10, TMSB4X, TPI1, TPM1, TPT1, TSPAN1, UBA52, UGT2B7, UMOD, UQCRB, UQCRFS1, VIM, WFDC2.
\end{flushleft}
\end{footnotesize}

We clustered the 200 genes into 31 groups. The group membership is listed below.

\begin{longtable}{p{0.06\linewidth} p{0.9\linewidth}}
\toprule
\textbf{Group} & \textbf{Members} \\
\midrule
0 & CLU, COX6B1, CRIM1, DCN, HLA-E, IGHA1, MGP, SAT1, SOD1, SPP1, TMA7, WFDC2 \\
1 & COX7B, RTN4 \\
2 & REN \\
3 & ADGRG1, ANPEP, ANXA2, APOE, MUC1, OAZ1, OGDHL, PODXL, S100A2, SOD2, SUCLG1, TSPAN1 \\
4 & A2M, ACADVL, ACTA2, ACTG1, AEBP1, AQP1, IGLC2, MALAT1, MIOX, MMP7, OST4, SFRP1 \\
5 & ALDOB, CALB1, CANX, CD81, NDRG1, NDUFB8, PIGR, PTH1R, RHOA, S100A10, SELENOP, UQCRB \\
6 & A1BG, APP, ASS1, CD24, DUSP1, EEF1D, HLA-DRA, HNRNPA1, NDUFA13, S100A6, UBA52, UQCRFS1 \\
7 & ACTB, ATP5MC3, HLA-DRB1, IGLC1, ITM2B, MAL, NME2, PGK1, PPP1R1A, RNASE1, SERPINA1, SLC12A1 \\
8 & ATP5F1D, CD151, CFL1, CKB, CST3, FTL, GSTP1, IDH2, IGFBP5, MGST3, NDUFB2, PDZK1IP1 \\
9 & ADIRF, ATP1B1, CLCNKB, COX5A, FLNA, HNRNPA2B1, HSD11B2, HTRA1, LAPTM4A, SLC25A3, SRP14, VIM \\
10 & ATP5MD, ATP5ME, BGN, CIRBP, DEFB1, GPX3, IFITM3, IGHG1, LGALS1, LUM, NDUFA1, PTGDS \\
11 & AQP2, BCAM, BSG, HINT1, KNG1, LAMP1, LRP2, NAT8, PEPD, SERPINA5, SLC25A6, SPARC \\
12 & ATP1A1, COL4A2, COX6C, DSTN, EPAS1, FTH1, IGFBP7, IGKC, SLC12A3, SLC3A1, TIMP1, UGT2B7 \\
13 & ATP5MF, B2M, CD74, EEF2, FXYD2, GATM, H3F3A, HLA-A, NDUFB7, P4HB, PFN1, TAGLN \\
14 & ATP5MPL, COL3A1, ENG, IGHG4, IGLC3, MYL9, NDUFA4, NEAT1, PEBP1, SLC13A3, TMSB10, TPI1 \\
15 & CA2, COX7A2, CTSB, EIF3K, HLA-B, IGFBP4, LDHA, PCK1, SPINK1, SSR4, TIMP3, TPM1 \\
16 & SLC5A12 \\
17 & COX7C \\
18 & CHCHD10, DYNLL1, GABARAP, HSPA8 \\
19 & CALM2 \\
20 & COX5B \\
21 & ATP6V0C \\
22 & UMOD \\
23 & NDUFB9 \\
24 & CRYAB \\
25 & HSPB1 \\
26 & TMSB4X \\
27 & C7 \\
28 & COL1A2, CTSH, CXCL14, CYSTM1, DDX5, EEF1G, EZR, HLA-C, IGHG3, LAMTOR5, MYL6, TPT1 \\
29 & SLC25A5 \\
30 & IFITM2 \\
\bottomrule
\end{longtable}

\paragraph{Examples of biologically coherent modules.}
Several groups align with well-known renal and tumor microenvironment programs:
\begin{itemize}
  \item \textbf{Thick ascending limb and distal nephron} (groups \textbf{22}, \textbf{7}, \textbf{12}): \textit{UMOD}, \textit{SLC12A1}, \textit{SLC12A3}. These are canonical markers of the thick ascending limb and distal convoluted tubule~\cite{devuyst2017umod,gamba2005slc12}.
  \item \textbf{Collecting duct principal cells} (group \textbf{11}): \textit{AQP2} with epithelial partners, consistent with vasopressin-regulated water transport~\cite{nielsen2002aqp2}.
  \item \textbf{Proximal tubule endocytosis and transport} (group \textbf{11}, \textbf{12}): \textit{LRP2} (megalin), \textit{SLC3A1}, consistent with proximal tubule uptake and amino acid handling~\cite{christensen2002megalin}.
  \item \textbf{Juxtaglomerular apparatus} (group \textbf{2}): \textit{REN} marks renin-producing cells~\cite{sequeira2015renin}.
  \item \textbf{Interstitial hypoxia and EPO axis} (group \textbf{12}): \textit{EPAS1} (HIF-2\(\alpha\)) associated with renal interstitial EPO-producing cells~\cite{kapitsinou2010hif2}.
  \item \textbf{Stromal ECM and smooth muscle/pericyte} (groups \textbf{4}, \textbf{13}, \textbf{14}, \textbf{28}): \textit{COL1A1/1A2/3A1}, \textit{DCN}, \textit{TAGLN}, \textit{MYL9}, typical of fibroblasts and mural cells~\cite{chen2021caf}.
  \item \textbf{Antigen presentation and interferon-stimulated genes} (groups \textbf{6}, \textbf{7}, \textbf{10}, \textbf{28}, \textbf{30}): \textit{HLA-DRA/DRB1/A/B/C}, \textit{CD74}, \textit{IFITM2/3} reflecting MHC and IFN-response modules~\cite{collins1984ifngHLA,schoggins2011isg}.
  \item \textbf{C1Q\textsuperscript{+} macrophages} (group \textbf{7}): presence of \textit{C1QA} with \textit{LGALS3} is consistent with C1Q\textsuperscript{+} TAM subsets~\cite{zhang2024tam}.
  \item \textbf{Injury-associated tubular program} (group \textbf{0}): \textit{SPP1} (osteopontin) often rises in stressed or injured tubules~\cite{liaw1998spp1}.
\end{itemize}

\subsection{Mouse Brain Dataset Selected Genes}

The following genes were selected for the Healthy Mouse Brain dataset:

\begin{footnotesize}
\begin{flushleft}
1110008P14Rik, 6330403K07Rik, Acot7, Apod, Apoe, App, Arpp19, Arpp21, Atp1a1, Atp1a2, Atp1b1, Atp2a2, Atp2b2, Atp5b, Atp5e, Atp5g1, Atp5j, Atp5o, Atp6v1e1, Baiap2, Basp1, Bc1, Bex2, Bsg, Calm2, Calm3, Camk2a, Camk2n1, Camkv, Cck, Cd81, Cfl1, Chgb, Chn1, Chst1, Ckb, Clstn1, Cox5a, Cox5b, Cox6a1, Cox6b1, Cox6c, Cox7b, Cox7c, Cox8a, Cplx1, Cplx2, Cryab, Cst3, Ctsd, Ctxn1, Dbi, Dclk1, Dnm1, Dpysl2, Dynll1, Dynll2, Eef1a1, Eno1, Eno2, Fkbp1a, Fkbp8, Fth1, Ftl1, Gaa, Gad1, Gas5, Gdi1, Gm42418, Gnao1, Gnas, Gng3, Gpi1, Gpm6a, Gpm6b, Gprasp1, Grcc10, H2afz, Hba-a1, Hba-a2, Hbb-bs, Hint1, Hpca, Hpcal4, Hspa8, Kif1a, Kif5a, Lars2, Ldhb, Lrrc17, Ly6h, Maged1, Malat1, Mbp, Mdh2, Meg3, Mlf2, Mobp, Mrfap1, Mt1, Myl12b, Myl6, Naca, Nap1l5, Ncdn, Ndfip1, Ndrg2, Ndufa12, Ndufa2, Ndufa3, Ndufa4, Ndufb9, Ndufc1, Nisch, Nnat, Nptxr, Nrgn, Nsf, Nsg1, Oaz1, Olfm1, Pcp4, Pde1b, Pea15a, Penk, Pfdn5, Pfn2, Pja2, Plp1, Ppp1r1b, Ppp3ca, Prkar1b, Ptgds, Ptma, Ptprn, Rab3a, Rnasek, Rpl10, Rpl13, Rpl13a, Rpl14, Rpl18, Rpl18a, Rpl19, Rpl22l1, Rpl23, Rpl23a, Rpl27, Rpl27a, Rpl29, Rpl32, Rpl34, Rpl36a, Rpl37, Rpl39, Rpl4, Rpl5, Rpl6, Rpl7, Rpl9, Rplp2, Rps12, Rps15a, Rps17, Rps18, Rps19, Rps2, Rps20, Rps23, Rps24, Rps27a, Rps3, Rps4x, Rps6, Rps7, Rtn3, Rtn4, Scd2, Scn1b, Selenow, Serf2, Sez6l2, Slc17a7, Slc1a2, Slc22a17, Slc25a4, Slc25a5, Snap25, Snap47, Snrpn, Sod1, Sparcl1, Sst, Stmn1, Stmn3, Sub1, Syn1, Syn2, Syngr1, Syt11, Tmsb10, Tmsb4x, Tpi1, Tspan7, Tuba1a, Tubb2a, Tubb4a, Tubb5, Ubl5, Uchl1, Uqcrc1, Uqcrh, Vsnl1, Wbp2, Ywhae, Ywhag, Ywhah.
\end{flushleft}
\end{footnotesize}

We clustered the 200 genes into 32 groups. The group membership is listed below.

\begin{longtable}{p{0.06\linewidth} p{0.9\linewidth}}
\toprule
\textbf{Group} & \textbf{Members} \\
\midrule
0 & Gnao1 \\
1 & Atp5b, Basp1, Camk2n1, Gas5, Nptxr, Pea15a, Rpl27a, Rpl29, Rps18, Rps20, Rps23, Stmn3 \\
2 & Stmn1 \\
3 & Cst3, Ctsd, Gad1, Gdi1, Mobp, Ndrg2, Ptprn, Rab3a, Rnasek, Rpl10, Rpl13a, Rpl19 \\
4 & Cplx1, Gnas, Gpm6a, Gprasp1, Hspa8, Kif1a, Ly6h, Nrgn, Penk, Pfdn5, Plp1, Ppp1r1b \\
5 & Arpp21, Cox8a, Malat1, Mbp, Mdh2, Meg3, Naca, Ndufa12, Ndufa2, Pja2, Rpl13, Rps6 \\
6 & Apod, Atp2b2, Bex2, Camkv, Rpl34, Rps19, Rps24, Rps27a, Rps4x, Rps7, Sez6l2, Snap25 \\
7 & 6330403K07Rik, Arpp19, Cox5a, Dpysl2, Ndufc1, Rpl18, Rpl39, Rpl7, Rpl9, Rps12, Rtn4, Slc1a2 \\
8 & Atp6v1e1, Dbi, Kif5a, Maged1, Nisch, Ppp3ca, Ptgds, Ptma, Rpl18a, Rplp2, Rps15a, Scd2 \\
9 & Apoe, Cfl1, Dynll2, Eno1, Hba-a2, Hbb-bs, Ldhb, Ndufb9, Rpl6, Rps2, Sub1, Syn1 \\
10 & Camk2a, Ckb, Cox6c, Fkbp8, H2afz, Hba-a1, Hpca, Lars2, Ndfip1, Rpl32, Sparcl1, Syt11 \\
11 & Atp5g1, Clstn1, Eef1a1, Fth1, Mrfap1, Myl12b, Nnat, Pcp4, Rpl36a, Rps3, Selenow, Sst \\
12 & 1110008P14Rik, Atp5o, Bc1, Chn1, Gpi1, Hpcal4, Lrrc17, Mlf2, Myl6, Ncdn, Pde1b, Snap47 \\
13 & App, Cck, Cryab, Dnm1, Dynll1, Hint1, Rpl22l1, Rpl23a, Slc25a4, Snrpn, Sod1, Syn2 \\
14 & Baiap2, Bsg, Calm3, Cplx2, Fkbp1a, Gaa, Gpm6b, Nsf, Rpl14, Rpl5, Rps17, Serf2 \\
15 & Acot7, Atp1a2, Atp5e, Calm2, Chgb, Chst1, Cox6a1, Eno2, Olfm1, Rpl23, Rpl37, Rpl4 \\
16 & Atp2a2, Cox7b, Ctxn1, Dclk1, Gm42418, Gng3, Oaz1, Rpl27, Rtn3, Scn1b, Slc17a7, Slc25a5 \\
17 & Cd81, Cox6b1, Prkar1b \\
18 & Atp5j \\
19 & Ndufa4 \\
20 & Atp1a1 \\
21 & Cox5b \\
22 & Mt1 \\
23 & Cox7c \\
24 & Nsg1 \\
25 & Atp1b1 \\
26 & Syngr1 \\
27 & Pfn2 \\
28 & Grcc10 \\
29 & Nap1l5 \\
30 & Slc22a17 \\
31 & Ndufa3 \\
\bottomrule
\end{longtable}

\paragraph{Examples of biologically coherent modules.}
Several groups align with well-known brain cell programs:
\begin{itemize}
  \item \textbf{Excitatory neurons} (groups \textbf{10}, \textbf{16}, \textbf{4}): \textit{Slc17a7} (VGLUT1), \textit{Camk2a}, \textit{Nrgn}, \textit{Hpca} mark glutamatergic neurons~\cite{tasic2018celltypes,yao2021m1atlas}.
  \item \textbf{Inhibitory neurons and neuropeptides} (groups \textbf{3}, \textbf{11}, \textbf{13}): \textit{Gad1}, \textit{Sst}, \textit{Cck} represent GABAergic interneuron classes~\cite{tasic2018celltypes,zeisel2015celltypes}.
  \item \textbf{Oligodendrocytes and myelination} (groups \textbf{5}, \textbf{4}, \textbf{3}): \textit{Mbp}, \textit{Plp1}, \textit{Mobp} are canonical myelin genes~\cite{cahoy2008transcriptome,marques2016oligo}.
  \item \textbf{Astrocyte-enriched genes} (groups \textbf{3}, \textbf{7}, \textbf{9}): \textit{Slc1a2}, \textit{Ndrg2}, \textit{Apoe} are enriched in astrocytes~\cite{zhang2014brainrna,srinivasan2016astro}.
  \item \textbf{Synaptic vesicle and release machinery} (groups \textbf{6}, \textbf{9}, \textbf{13}): \textit{Snap25}, \textit{Syn1}/\textit{Syn2}, \textit{Syt11}, \textit{Rab3a}, \textit{Dnm1} participate in synaptic transmission~\cite{sudhof2013lastms}.
\end{itemize}

\subsection{PRAD Dataset Selected Genes}

The following genes were selected for the PRAD (prostate cancer) dataset:

\begin{footnotesize}
\begin{flushleft}
A2M, ACPP, ACTA2, ACTB, ACTG1, ACTG2, ADIRF, AGR2, AMD1, APLP2, ATP5F1E, ATP5IF1, ATP5MD, ATP5MF, ATP5MPL, ATP6V0B, AZGP1, B2M, BTF3, C12orf57, CALM2, CALR, CD63, CD74, CD81, CD9, CD99, CFD, CFL1, CHCHD2, CIRBP, CKB, CLU, CNN1, COMMD6, COPS9, COX4I1, COX5B, COX6A1, COX6C, COX7A2, COX7B, COX7C, COX8A, CPE, CSRP1, CST3, DBI, DDT, DES, DHRS7, DSTN, DUSP1, EDF1, EEF1A1, EEF1B2, EEF2, EGR1, EIF1, EIF3L, ELOB, FABP5, FASN, FAU, FBLN1, FLNA, FOS, FTH1, FTL, FXYD3, GAPDH, GPX4, H2AFJ, H3F3A, H3F3B, HERPUD1, HINT1, HLA-B, HLA-C, HLA-DRA, HMGN2, HNRNPA1, HOXB13, HSPA5, HSPA8, HSPB1, IGHA1, IGKC, IGLC2, ITM2B, KDELR2, KLK2, KLK3, KLK4, KRT18, KRT8, LGALS1, LTF, MALAT1, MDK, MGP, MIF, MINOS1, MPC2, MSMB, MYH11, MYL6, MYL9, MYLK, MZT2B, NACA, NBL1, NDRG1, NDUFB1, NDUFB11, NDUFB4, NDUFS5, NEAT1, NEFH, NKX3-1, NME4, NPM1, NPY, NR4A1, NUPR1, OAZ1, OST4, PABPC1, PARK7, PDLIM5, PFDN5, PFN1, PLA2G2A, PLPP1, PMEPA1, POLR2L, PPDPF, PPIA, PRAC1, PRDX2, PRDX6, PTGDS, PTMA, RACK1, RDH11, ROMO1, RPN2, S100A11, S100A6, SARAF, SAT1, SEC11C, SEC61B, SEC61G, SELENOP, SELENOW, SERF2, SERP1, SKP1, SLC25A6, SLC45A3, SNHG19, SNHG25, SNHG8, SNRPD2, SORD, SPDEF, SPINT2, SPON2, SRP14, SSR4, STEAP2, TAGLN, TFF3, TIMP1, TMBIM6, TMEM141, TMEM258, TMEM59, TMPRSS2, TMSB10, TMSB4X, TOMM7, TPM2, TPT1, TRPM4, TSC22D3, TSPAN1, TSTD1, TXN, UBA52, UBB, UBL5, UQCR10, UQCRB, UQCRH, UQCRQ, VEGFA, VIM, ZFAS1.

We clustered the 200 genes into 32 groups. The group membership is listed below.

\begin{longtable}{p{0.06\linewidth} p{0.9\linewidth}}
\toprule
\textbf{Group} & \textbf{Members} \\
\midrule
0 & EDF1, EEF1B2, FABP5, HLA\textendash C \\
1 & SELENOP \\
2 & HERPUD1 \\
3 & A2M, CFL1, CIRBP, PPDPF, PRDX2, PRDX6, PTGDS, RACK1, RDH11, RPN2, S100A6, SARAF \\
4 & LTF, MYL6, MYLK, MZT2B, NACA, NBL1, NDRG1, OAZ1, ROMO1, SLC25A6, SORD, TMEM141 \\
5 & ACPP, ACTG2, ADIRF, AGR2, AMD1, ATP5F1E, BTF3, NDUFB4, TMBIM6, TMPRSS2, TMSB10, UQCRQ \\
6 & ACTG1, APLP2, COPS9, EIF3L, NDUFS5, NUPR1, OST4, PRAC1, SAT1, TMSB4X, TOMM7, TSPAN1 \\
7 & ACTA2, AZGP1, B2M, CNN1, EIF1, HSPA5, HSPB1, LGALS1, NME4, PMEPA1, S100A11, UQCRH \\
8 & ATP5IF1, COX7C, HOXB13, IGKC, KDELR2, MDK, NEFH, SKP1, TPM2, TPT1, UBB, ZFAS1 \\
9 & ATP5MF, COX4I1, ELOB, HMGN2, HSPA8, MPC2, PLPP1, POLR2L, PPIA, SRP14, TMEM258, TSTD1 \\
10 & CD74, CKB, COX7B, CPE, DHRS7, EEF2, HLA\textendash DRA, PDLIM5, SELENOW, TXN, UBL5, UQCRB \\
11 & ACTB, ATP6V0B, CD81, CD9, FLNA, FTH1, H3F3A, HNRNPA1, IGLC2, MYL9, NR4A1, SNHG25 \\
12 & ATP5MPL, CD63, COX5B, FTL, MGP, NDUFB1, NDUFB11, PFN1, PTMA, SEC61G, SNHG8, SNRPD2 \\
13 & ATP5MD, C12orf57, COMMD6, CSRP1, FASN, FXYD3, KLK3, MIF, NEAT1, SEC11C, SPINT2, VEGFA \\
14 & CD99, CLU, DBI, FBLN1, KRT8, MALAT1, NPY, PLA2G2A, SEC61B, SERF2, SPDEF, TAGLN \\
15 & COX6A1, COX7A2, EGR1, FAU, H3F3B, MSMB, MYH11, NKX3\textendash 1, PFDN5, SLC45A3, TFF3, TIMP1 \\
16 & CALM2, CHCHD2, CST3, DES, GAPDH, H2AFJ, HINT1, IGHA1, KLK2, SNHG19, SPON2, SSR4 \\
17 & CALR, CFD, COX8A, DSTN, GPX4, KRT18, PABPC1, PARK7, SERP1, STEAP2, TSC22D3 \\
18 & TMEM59 \\
19 & VIM \\
20 & EEF1A1 \\
21 & KLK4 \\
22 & DUSP1 \\
23 & ITM2B \\
24 & MINOS1 \\
25 & UQCR10 \\
26 & NPM1 \\
27 & DDT \\
28 & HLA\textendash B \\
29 & COX6C \\
30 & TRPM4 \\
31 & FOS \\
\bottomrule
\end{longtable}

\paragraph{Examples of biologically coherent modules.}
Several groups align with well-known PRAD biology:
\begin{itemize}
  \item \textbf{Androgen\textendash regulated luminal secretory program} (groups \textbf{13}, \textbf{15}, \textbf{8}, \textbf{5}, \textbf{16}, \textbf{21}): \textit{KLK3}/\textit{KLK2}, \textit{NKX3\textendash 1}, \textit{HOXB13}, \textit{TMPRSS2}, \textit{SLC45A3}, \textit{MSMB}, \textit{TFF3}, \textit{SPDEF}. These genes are lineage markers or direct androgen receptor targets in prostate epithelium~\cite{cleutjens1997psa,wang2009arcistrome,ewing2012hoxb13,tomlins2005tmprss2}.
  \item \textbf{Stromal smooth muscle and CAF\textendash like ECM} (groups \textbf{7}, \textbf{14}, \textbf{15}): \textit{ACTA2}, \textit{CNN1}, \textit{MYH11}, \textit{TAGLN}, \textit{FBLN1}, \textit{DES} mark prostate stroma and myofibroblasts~\cite{chen2021caf}.
  \item \textbf{Antigen presentation and immunoglobulin} (groups \textbf{10}, \textbf{28}, \textbf{11}, \textbf{8}, \textbf{16}): \textit{HLA\textendash DRA}/\textit{HLA\textendash B}, \textit{CD74}, \textit{IGKC}/\textit{IGLC2}/\textit{IGHA1} reflect MHC and B\textendash cell modules~\cite{collins1984ifngHLA}.
  \item \textbf{Prostate\textendash enriched antigens and secreted factors} (groups \textbf{7}, \textbf{17}, \textbf{16}): \textit{AZGP1}, \textit{STEAP2}, \textit{SPON2} are well\textendash documented prostate\textendash enriched proteins with diagnostic or biological relevance~\cite{hubert1999steap2,zhang2012spon2}.
\end{itemize}

\subsection{ccRCC Dataset Selected Genes}

The following genes were selected for the ccRCC (clear cell Renal Cell Carcinoma) dataset:

\begin{footnotesize}
\begin{flushleft}
A2M, ACTA2, ACTB, ACTR3, ADIRF, AEBP1, AHNAK, ANGPTL4, ANPEP, ANXA2, APOC1, APOE, APOL1, APP, ARF1, ARF4, ARPC1B, ARPC2, ASPH, ATP1A1, ATP1B1, ATP5F1B, ATP5MC2, ATP5ME, B2M, BGN, BIRC3, BRI3, BSG, BST2, C19orf33, C1QA, C1QB, C1QC, C1R, C1S, C3, CA12, CALD1, CANX, CAV1, CCDC91, CCN1, CCN2, CCNI, CD24, CD44, CD63, CD68, CD74, CD81, CD99, CEBPD, CHCHD2, CIRBP, CLU, COL18A1, COL1A1, COL1A2, COL3A1, COL4A1, COL4A2, COL6A1, COL6A2, COL6A3, COX6C, CP, CPE, CRYAB, CST3, CSTB, CTSA, CTSB, CTSD, CTSZ, CXCR4, CYB5A, CYB5R3, DCN, DDIT4, DDX17, DEPP1, DUSP1, EEF1G, EIF1, EIF4A1, EIF4A2, EIF4G2, EIF4H, ENPP3, FCGRT, FGB, FKBP5, FLNA, FN1, FOS, FTH1, FTL, FXYD2, GABARAP, GLUL, GPX3, GSN, GSTP1, H3F3B, HINT1, HIST1H4C, HLA-A, HLA-DPA1, HLA-DPB1, HLA-DQA1, HLA-DQB1, HLA-DRA, HLA-DRB1, HLA-F, HMGB1, HNRNPA2B1, HNRNPA3, HPCAL1, HSP90AA1, HSP90B1, HSPA1B, HSPA5, HSPA8, HSPD1, HSPG2, HTRA1, IFI27, IFI30, IFI6, IFITM2, IGFBP3, IGFBP4, IGFBP5, IGFBP7, IGHA1, IGHG1, IGHG2, IGHG3, IGHG4, IGHM, IGKC, IGLC1, IL32, ITGA3, ITGB1, JCHAIN, KRT18, KRT8, LAPTM5, LGALS1, LGALS3, LGALS3BP, LY6E, LYZ, MCL1, MGP, MIF, MMP7, MYH9, MYL9, NCL, NDRG1, NDUFA4L2, NME2, NOP53, NPC2, NUPR1, P4HB, PARK7, PCBP1, PCBP2, PDIA6, PDK4, PDZK1IP1, PEBP1, PFDN5, PFKP, PGAM1, PGF, PLEC, PLIN2, PLOD2, PLTP, POSTN, PPDPF, PPP2CB, PRDX6, PRR13, PSAP, PTMA, PTMS, PTTG1IP, RARRES2, RASSF4, RGS5, RHOB, RNASET2, RPN2, S100A11, S100A6, SAT1, SCD, SEC61G, SELENOP, SERPINA1, SERPINE1, SERPING1, SNX3, SOD2, SPARC, SPINK13, SPP1, SQSTM1, SRRM2, SRSF2, SSR4, TAGLN, TAGLN2, TGFBI, TGM2, THBS1, TIMP1, TIMP3, TMBIM6, TMEM176A, TMEM176B, TMSB4X, TOMM7, TPM1, TPM2, TRAM1, TSC22D3, TUBA1B, TXN, TXNIP, TYROBP, UBA52, UBC, UQCRQ, VEGFA, VIM, VWF.
\end{flushleft}
\end{footnotesize}

We clustered the 200 genes into 32 groups. The group membership is listed below.

\begin{longtable}{p{0.06\linewidth} p{0.9\linewidth}}
\toprule
\textbf{Group} & \textbf{Members} \\
\midrule
0 & ANGPTL4, APP, ATP1B1, BGN, BIRC3, CAV1, CD74, FTH1, FTL, GABARAP, HNRNPA3, PTMA \\
1 & GLUL \\
2 & CYB5A \\
3 & ARPC1B, C1R, CA12, CD63, CIRBP, COL4A2, DDIT4, FGB, HSP90B1, HTRA1, PDZK1IP1, RNASET2 \\
4 & AEBP1, AHNAK, FKBP5, GSTP1, HIST1H4C, HLA\textendash DQB1, HMGB1, HSPG2, IGFBP4, ITGB1, NDUFA4L2, PSAP \\
5 & BST2, HLA\textendash DQA1, HLA\textendash DRA, HNRNPA2B1 \\
6 & NUPR1 \\
7 & KRT18 \\
8 & CCNI \\
9 & COL4A1 \\
10 & C1QB \\
11 & IGKC \\
12 & ANPEP, COX6C, HSPA8, IFI30, IGHG4, IGHM, LY6E, MCL1, MGP, MYL9, NCL, PARK7 \\
13 & APOC1, GSN, HSPD1, IFITM2, KRT8, LAPTM5, LGALS1, LGALS3, MMP7, MYH9, PLIN2, PTMS \\
14 & ADIRF, ARF4, CCDC91, COL18A1, CTSZ, FXYD2, HLA\textendash A, HSP90AA1, ITGA3, LYZ, NPC2, PEBP1 \\
15 & APOE, ATP5MC2, BRI3, BSG, C1S, CP, DDX17, DEPP1, GPX3, IGFBP3, LGALS3BP, PGAM1 \\
16 & ACTB, ARPC2, C1QA, C1QC, CCN1, COL6A3, CYB5R3, IGFBP5, MIF, PFDN5, PPDPF, S100A11 \\
17 & A2M, ACTR3, ANXA2, APOL1, ATP1A1, HLA\textendash DPA1, HSPA1B, HSPA5, NDRG1, PLTP, RARRES2, RGS5 \\
18 & ATP5F1B, CD81, CST3, CXCR4, EIF4H, ENPP3, FCGRT, FN1, IGHG2, NOP53, PRR13, RPN2 \\
19 & CALD1, CTSA, CTSB, IFI6, IGHA1, IGHG3, NME2, PCBP2, PDIA6, POSTN, PRDX6, RHOB \\
20 & ARF1, CCN2, CD24, CD44, COL6A2, EIF4G2, IFI27, IGHG1, JCHAIN, P4HB, PGF, PLOD2 \\
21 & ACTA2, ASPH, C3, CANX, CD99, CEBPD, COL1A1, DCN, EIF4A1, FOS, HLA\textendash DPB1, HPCAL1 \\
22 & ATP5ME, CLU, CPE, CSTB, DUSP1, EEF1G, EIF1, FLNA, HLA\textendash F, IGLC1, PLEC, RASSF4 \\
23 & C19orf33, CD68, CHCHD2, COL1A2, COL3A1, COL6A1, CRYAB, H3F3B, HINT1, IL32, PDK4, PPP2CB \\
24 & CTSD \\
25 & PTTG1IP \\
26 & EIF4A2 \\
27 & PFKP \\
28 & HLA\textendash DRB1 \\
29 & IGFBP7 \\
30 & B2M \\
31 & PCBP1 \\
\bottomrule
\end{longtable}

\paragraph{Examples of biologically coherent modules (by groups).}
Several groups align with well-known ccRCC or tumor-microenvironment programs:
\begin{itemize}
  \item \textbf{Hypoxia/lipid metabolism and ECM remodeling}: groups \textbf{3}, \textbf{4}, \textbf{15}, \textbf{18}, \textbf{20}. Representative genes include \textit{NDUFA4L2}, \textit{PLIN2}, \textit{CA12}, \textit{ENPP3}, \textit{PLOD2}, \textit{PGF}, \textit{FN1}. These are classic HIF–hypoxia targets or matrix/secretory programs frequently upregulated in ccRCC~\cite{kubala2023ndufa4l2,cao2018plin2,ivanov2001ca12,thompson2018enpp3}.
  \item \textbf{Antigen presentation and interferon-stimulated genes}: groups \textbf{5}, \textbf{14}, \textbf{21}, \textbf{28}. Genes such as \textit{HLA\textendash DRA/DRB1/DPA1/DPB1}, \textit{HLA\textendash A}, \textit{BST2}, \textit{IFI27}, \textit{IFI6}, \textit{IFITM2}, \textit{LY6E} mark MHC-II antigen processing and IFN response~\cite{collins1984ifngHLA,ortega2024isg}.
  \item \textbf{C1Q\textsuperscript{+} macrophage module}: groups \textbf{10}, \textbf{16}, \textbf{23} with \textit{C1QA/B/C}, \textit{CD68}, often alongside \textit{APOE}/\textit{LGALS3}~\cite{lindblom2003rgs5pericyte,he2024mural}.
  \item \textbf{Pericyte/smooth-muscle and stromal ECM}: groups \textbf{19}, \textbf{21}, \textbf{23} with \textit{ACTA2}, \textit{MYL9}, \textit{CALD1}, \textit{RGS5}, and collagens \textit{COL1A1/1A2/3A1/6A1/6A3}, \textit{DCN}, \textit{POSTN}, \textit{FN1}~\cite{lindblom2003rgs5pericyte,he2024mural}.
  \item \textbf{Plasma-cell immunoglobulins}: groups \textbf{11}, \textbf{12}, \textbf{18}, \textbf{19}, \textbf{20} featuring \textit{IGHG1/2/3/4}, \textit{IGHM}, \textit{IGHA1}, \textit{IGKC}, \textit{JCHAIN}~\cite{xu2020jchain,onieva2022igkc}.
\end{itemize}

\section{Implementation Details and Reproducibility}
\mypara{Reproducibility}
To ensure consistent results, we fix the random seed at 2021 (configurable via \texttt{--seed}). The \texttt{fix\_seed} function controls randomness in Python, NumPy, PyTorch, and CUDA operations. For complete reproducibility, set Lightning's \texttt{deterministic=True} and apply cuDNN flags as recommended in PyTorch documentation. All inference scripts use fixed seeds.

\mypara{Training Setup}
We use a learning rate of 1e-4 with weight decay of 1e-4 and gradient clipping at 1.0. The learning rate scheduler is disabled by default. Training uses batch size 256 while validation and testing use batch size 64, with 4 data loading workers and \texttt{pin\_memory=True}.

The model uses 768-dimensional embeddings with 8 transformer layers, 8 attention heads, and MLP ratio of 3.0. Dropout is set to 0.0 for training and 0.1 for evaluation. Multi-scale patches are configured as \texttt{gene\_patch\_nums = (1, 4, 8, 40, 100, 200)} with a vocabulary size of \texttt{max\_gene\_count + 1} (default 2000). All models train for 50 epochs with early stopping.

\mypara{Multi-GPU Training}
We use DDP for multi-GPU training with \texttt{accumulate\_grad\_batches = 1} and \texttt{find\_unused\_parameters = False}. DDP is automatically enabled when multiple GPUs are detected.

\section{Additional Visualization Results}

To provide comprehensive spatial visualization comparisons across different genes, we present additional visualization results on the HER2ST dataset using the SPA148 sample. These visualizations demonstrate the spatial expression patterns predicted by our GenAR method compared to baseline approaches across a diverse set of genes with different expression characteristics and biological functions. The selected genes represent various functional categories including structural proteins, growth factors, immune-related genes, and metabolic enzymes, showcasing the generalization capability of our method across different gene types.

\section{Large Language Model Usage}

Large Language Models were used as general-purpose writing assistance tools to improve the grammar, clarity, and organisation of the manuscript. The core research contributions, methodology, experimental design, and scientific insights are entirely original work by the authors.

\newpage
\begin{figure*}[t]
\centering
\includegraphics[width=1\textwidth]{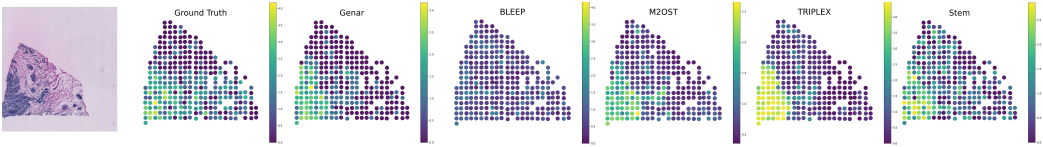}
\caption{Spatial visualization comparison of C12orf57 gene expression prediction on HER2ST SPA148 sample.}
\label{fig:vis_c12orf57}
\end{figure*}

\begin{figure*}[t]
\centering
\includegraphics[width=1\textwidth]{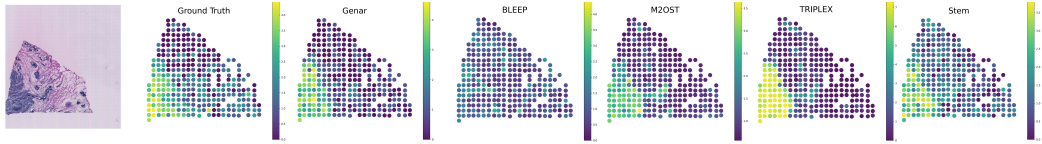}
\caption{Spatial visualization comparison of EIF4G1 gene expression prediction on HER2ST SPA148 sample.}
\label{fig:vis_eif4g1}
\end{figure*}

\begin{figure*}[t]
\centering
\includegraphics[width=1\textwidth]{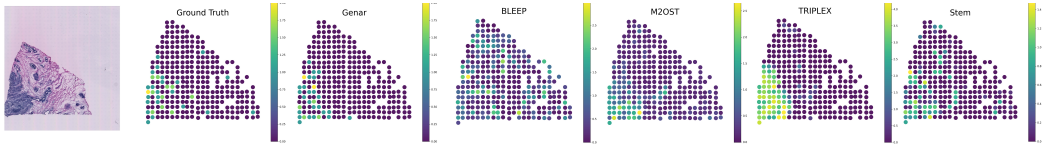}
\caption{Spatial visualization comparison of FNBP1L gene expression prediction on HER2ST SPA148 sample.}
\label{fig:vis_fnbp1l}
\end{figure*}

\begin{figure*}[t]
\centering
\includegraphics[width=1\textwidth]{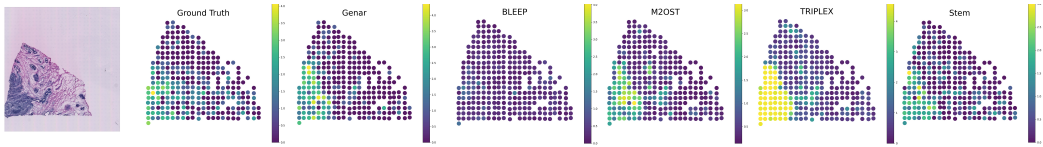}
\caption{Spatial visualization comparison of IGFBP2 gene expression prediction on HER2ST SPA148 sample.}
\label{fig:vis_igfbp2}
\end{figure*}

\begin{figure*}[t]
\centering
\includegraphics[width=1\textwidth]{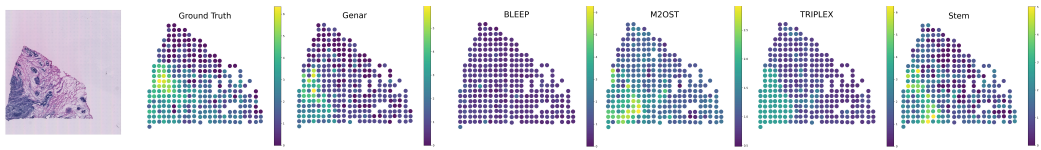}
\caption{Spatial visualization comparison of ISG15 gene expression prediction on HER2ST SPA148 sample.}
\label{fig:vis_isg15}
\end{figure*}

\clearpage

\begin{figure*}[t]
\centering
\includegraphics[width=1\textwidth]{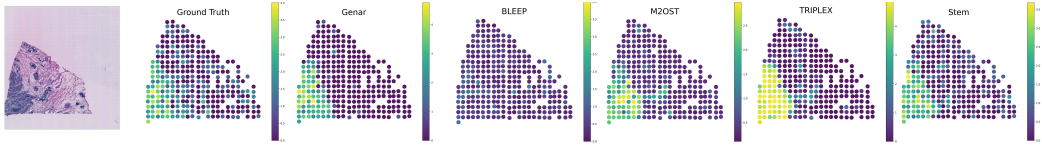}
\caption{Spatial visualization comparison of NUCKS1 gene expression prediction on HER2ST SPA148 sample.}
\label{fig:vis_nucks1}
\end{figure*}

\begin{figure*}[t]
\centering
\includegraphics[width=1\textwidth]{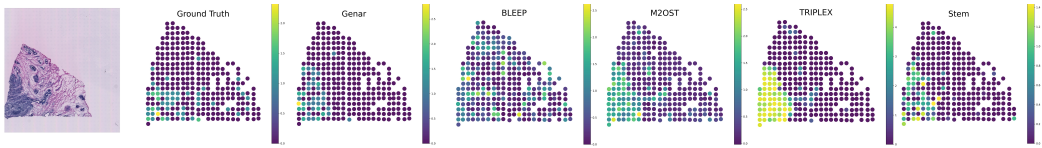}
\caption{Spatial visualization comparison of ORMDL3 gene expression prediction on HER2ST SPA148 sample.}
\label{fig:vis_ormdl3}
\end{figure*}

\begin{figure*}[t]
\centering
\includegraphics[width=1\textwidth]{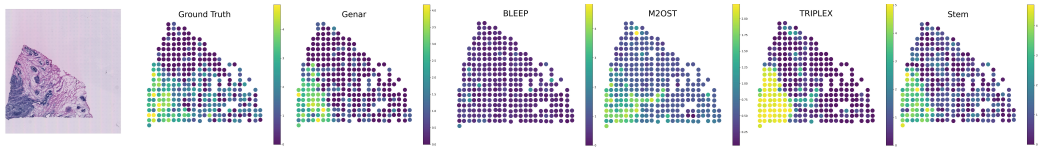}
\caption{Spatial visualization comparison of PPP1R1B gene expression prediction on HER2ST SPA148 sample.}
\label{fig:vis_ppp1r1b}
\end{figure*}

\begin{figure*}[t]
\centering
\includegraphics[width=1\textwidth]{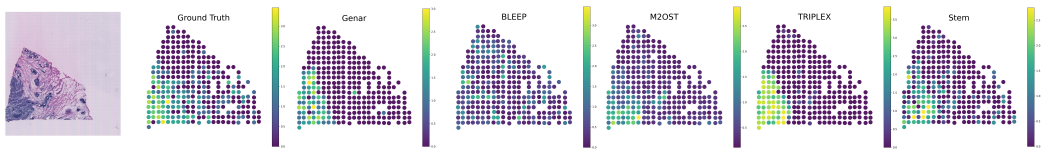}
\caption{Spatial visualization comparison of SF3B5 gene expression prediction on HER2ST SPA148 sample.}
\label{fig:vis_sf3b5}
\end{figure*}

\begin{figure*}[t]
\centering
\includegraphics[width=1\textwidth]{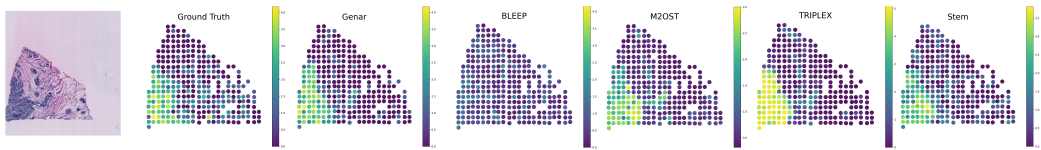}
\caption{Spatial visualization comparison of SSR2 gene expression prediction on HER2ST SPA148 sample.}
\label{fig:vis_ssr2}
\end{figure*}

\end{flushleft}
\end{footnotesize}
\end{document}